\documentclass[times, twocolumn, final]{elsarticle}

\usepackage{ycviu}
\usepackage{framed,multirow}

\usepackage{amssymb}
\usepackage{latexsym}
\usepackage{environ}
\makeatletter
\newsavebox{\measure@tikzpicture}
\NewEnviron{scaletikzpicturetowidth}[1]{%
  \def\tikz@width{#1}%
  \def\tikzscale{1}\begin{lrbox}{\measure@tikzpicture}%
  \BODY
  \end{lrbox}%
  \pgfmathparse{#1/\wd\measure@tikzpicture}%
  \edef\tikzscale{\pgfmathresult}%
  \BODY
}
\makeatother

\usepackage{pgfplots}
\pgfplotsset{compat=1.7}
\usepgfplotslibrary{groupplots}

\usepackage{xcolor}
\usepackage{subcaption}
\usepackage{booktabs}
\usepackage{multirow}

\usepackage{import}

\usepackage{url}
\usepackage{xcolor}
\definecolor{newcolor}{rgb}{.8,.349,.1}

\journal{Computer Vision and Image Understanding}

\begin{document}




\begin{frontmatter}

\title{Report on UG$^2+$ Challenge Track 1: Assessing  Algorithms to Improve Video Object Detection and Classification from Unconstrained Mobility Platforms}
\author[1,3]{Sreya \snm{Banerjee}\corref{cor1}} 
\cortext[cor1]{Corresponding author:} 
\ead{sbanerj2@nd.edu}
\author[1,3]{Rosaura G. \snm{VidalMata}}
\author[2]{Zhangyang \snm{Wang}}
\author[1]{Walter J. \snm{Scheirer}}
\address[1]{Dept. of Computer Science \& Engineering, University of Notre Dame, Notre Dame, IN, 46556, USA.}
\address[2]{Visual Informatics Group, University of Texas at Austin, Austin, TX, USA.}
\address[3]{denotes equal contribution}

\begin{abstract}
How can we effectively engineer a computer vision system that is able to interpret videos from unconstrained mobility platforms like UAVs? One promising option is to make use of image restoration and enhancement algorithms from the area of computational photography to improve the quality of the underlying frames in a way that also improves automatic visual recognition. Along these lines, exploratory work is needed to find out which image pre-processing algorithms, in combination with the strongest features and supervised machine learning approaches, are good candidates for difficult scenarios like motion blur, weather, and mis-focus --- all common artifacts in UAV acquired images. This paper summarizes the protocols and results of Track 1 of the UG$^2+$ Challenge held in conjunction with IEEE/CVF CVPR 2019. The challenge looked at two separate problems: (1) object detection improvement in video, and (2) object classification improvement in video. The challenge made use of new protocols for the UG$^2$ (UAV, Glider, Ground) dataset, which is an established benchmark for assessing the interplay between image restoration and enhancement and visual recognition. In total, $16$ algorithms were submitted by academic and corporate teams, and a detailed analysis of them is reported here.
\end{abstract}


\end{frontmatter}

\vspace{-1mm}
\section{Introduction}
\vspace{-2mm}
The use of mobile video capturing devices in unconstrained scenarios offers clear advantages in a variety of areas where autonomy is just beginning to be deployed. For instance, a camera installed on a platform like an unmanned aerial vehicle (UAV) could provide valuable information about a disaster zone without endangering human lives. And a flock of such devices can facilitate the prompt identification of dangerous hazards as well as the location of survivors, the mapping of terrain, and much more. However, the abundance of frames captured in a single session makes the automation of their analysis a necessity.

To do this, one's first inclination might be to turn to the state-of-the-art visual recognition systems which, trained with millions of images crawled from the web, would be able to identify objects, events, and human identities from a massive pool of irrelevant frames. However, such approaches do not take into account the artifacts unique to the operation of the sensors used to capture outdoor data, as well as the visual aberrations that are a product of the environment. While there have been important advances in the area of computational photography~\citep{ledig2017photo,Nah_2017_CVPR}, their incorporation as a pre-processing step for higher-level tasks has received only limited attention over the past few years \citep{Sharma2017,vidal2018ug}. The impact many transformations have on visual recognition algorithms remains unknown.

Following the success of the UG$^2$ Challenge on this topic held at IEEE/CVF CVPR 2018~\citep{vidal2018ug, DBLP:journals/corr/abs-1901-09482}, a new challenge with an emphasis on video was organized at CVPR 2019. The UG$^2+$ 2019 Challenge provided an integrated forum for researchers to evaluate recent progress in handling various adverse visual conditions in real-world scenes, in robust, effective and task-oriented ways.

$16$ novel algorithms were submitted by academic and corporate teams from the University of the Chinese Academy of Sciences (UCAS), Northeastern University (NEU), Institute of Microelectronics of the Chinese Academy of Sciences (IMECAS), University of Macau (UMAC), Honeywell International, Technical University of Munich (TUM), Chinese Academy of Sciences (CAS), Sunway.AI, and Meitu's MTlab.

In this paper, we review the changes made to the original dataset, evaluation protocols, algorithms submitted, and experimental analysis for Track $1$ of the challenge, primarily concerned with video object detection and classification from unconstrained mobility platforms. (A separate paper was published describing Track 2~\cite{yang2020advancing}, which focused on improving poor visibility environments.) The novelty of this work lies in the evaluation protocols we use to assess algorithms, which quantify the mutual influence between low-level computational photography approaches and high-level tasks such as detection and classification. Moreover, it is the first such work to rigorously evaluate video object detection and classification after task-specific image pre-processing. 

\vspace{-1mm}
\section{Related Work}
\vspace{-2mm}
\textbf{Datasets.} There is an ample number of datasets designed for the qualitative evaluation of image enhancement algorithms in the area of computational photography. Such datasets are often designed to fix a particular type of aberration such as blur~\citep{Levin:2009, kohler2012recording, sun2013edge, shi2014discriminative, Nah_2017_CVPR}, noise~\citep{abdelhamed2018high, chen2018learning, plotz2017benchmarking}, or low resolution~\citep{Agustsson_2017_CVPR_Workshops}. Datasets containing more diverse scenarios~\citep{huang2015single, sheikh2006statistical, yang2014single, Su:2016:DBN} have also been proposed. However, these datasets were designed for image quality assessment purposes, rather than for a quantitative evaluation of the enhancement algorithm on a higher-level task like recognition.

Datasets with a similar type of data to the one employed for this challenge include large-scale video surveillance datasets such as~\citep{CAVIAR:Dataset:2004, grgic2011scface, CUHK:Dataset:2014, TISI:Dataset:2013}, which provide video captured from a single fixed overhead viewpoint. As for datasets collected by aerial vehicles, the VIRAT~\citep{Virat:Dataset:2011} and VisDrone2018~\citep{zhuvisdrone2018} datasets have been designed for event recognition and object detection respectively. Other aerial datasets include the UCF Aerial Action Data Set~\citep{UCFAA}, UCF-ARG~\citep{UCFARG}, UAV123~\citep{mueller2016benchmark}, and the multi-purpose dataset introduced by Yao \emph{et al.}~\citep{LHI:Dataset:2007}. Similarly, none of these datasets provide protocols that introduce image enhancement techniques to improve the performance of visual recognition.

\textbf{Restoration and Enhancement to Improve Visual Recognition.}

Intuitively, improving the visual quality of a corrupted image should, in turn, improve the performance of object recognition for a classifier analyzing the image. As such, one could assume a correlation between the perceptual quality of an image and its quality for object recognition purposes, as has been observed by Gondal \emph{et al.} \citep{2018arXiv180800043W} and Tahboub \emph{et al.} \citep{8297071}. 

Early attempts at unifying visual recognition and visual enhancement tasks included deblurring~\citep{zeiler2010deconvolutional, zeiler2011adaptive}, super-resolution~\citep{DBLP:journals/corr/abs-1803-11316}, denoising~\citep{DBLP:journals/corr/abs-1710-06805}, and dehazing~\citep{8237773}. These approaches tend to overlook the qualitative appearance of the images and instead focus on improving the performance of object recognition. In contrast, the approach proposed by Sharma \emph{et al.}~\citep{Sharma2017} incorporates two loss functions for enhancement and classification into an end-to-end processing and classification pipeline. 

Outside of object recognition, visual enhancement techniques have been of interest for unconstrained face recognition~\citep{yao2008improving, nishiyama2009facial, zhang2011close, 4401949,DBLP:journals/corr/WuDXC16, lin2005face,  yu2011face, huang2011super, uiboupin2016facial, rasti2016convolutional, jing2015super} through the incorporation of deblurring, super-resolution, hallucination techniques, and person re-identification for video surveillance data. 

\vspace{-1mm}
\section{The UG$^2$+ Challenge} \label{Challenge}
\vspace{-2mm}
The main goal of this work is to provide insights related to the impact image restoration and enhancement techniques have on visual recognition tasks performed on video captured in unconstrained scenarios. For this, we introduce two visual recognition tasks: (1) \textit{object detection improvement in video}, where algorithms produce enhanced images to improve the localization and identification of an object of interest within a frame, and (2) \textit{object classification improvement in video}, where algorithms analyze a group of consecutive frames in order to create a better video sequence to improve classification of a given object of interest within those frames.

\vspace{-1mm}
\subsection{Object Detection Improvement in Video}\vspace{-0.5mm}
For Track 1.1 of the challenge, the UG$^2$ dataset~\citep{vidal2018ug} was adapted to be used for localizing and identifying objects of interest\footnote{The object detection dataset (including the train-validation split) and evaluation kit is available from: http://bit.ly/UG2Detection}. This new dataset exceeds PASCAL VOC~\citep{everingham2010pascal} in terms of the number of classes used, as well as in the difficulty of recognizing some classes due to image artifacts. 
$93,668$ object-level annotations were extracted from $195$ videos coming from the three original UG$^2$ collections~\citep{vidal2018ug} (UAV, Glider, Ground), spanning $46$ classes inspired by ImageNet~\citep{russakovsky2015imagenet}. There are $86,484$ video frames, each having a corresponding annotation file in .xml format, similar to PASCAL VOC. 

Each annotation file includes the dataset collection the image frame belongs to, its relative path, width, height, depth, objects present in the image, the bounding box coordinates indicating the location of each object in the image, and segmentation and difficulty indicators. (Note that different videos have different resolutions.) Since we are primarily concerned with localizing and recognizing the object, the indicator for segmentation in the annotation file is kept at $0$ meaning ``no segmentation data available." Because the objects in our dataset are fairly recognizable to humans, we kept the indicator for difficulty set to $0$ to indicate ``easy." 

Similar to the original UG$^2$ dataset, the UG$^2$+ object detection dataset is divided into the following three categories: (1) 30 Creative Commons tagged videos taken by fixed-wing UAVs obtained from YouTube; (2) $29$ glider videos recorded by pilots of fixed-wing gliders; and (3) $136$ controlled videos captured on the ground using handheld cameras. Unlike the original UG$^2$ dataset, we do not crop out the objects from the frames, and instead use the whole frames for the detection task. 

Common artifacts among all the three categories include glare/lens flare, poor image quality, occlusion, over/under exposure, camera shaking, noise, motion blur, fish eye lens distortion, and problematic weather conditions. The top row of Table~1 provides the training dataset statistics. Further details about the training dataset can be found in Supp. Sec 1.1.

\begin{figure}[hb!]
    \centering
    \includegraphics[width=1.0\linewidth]{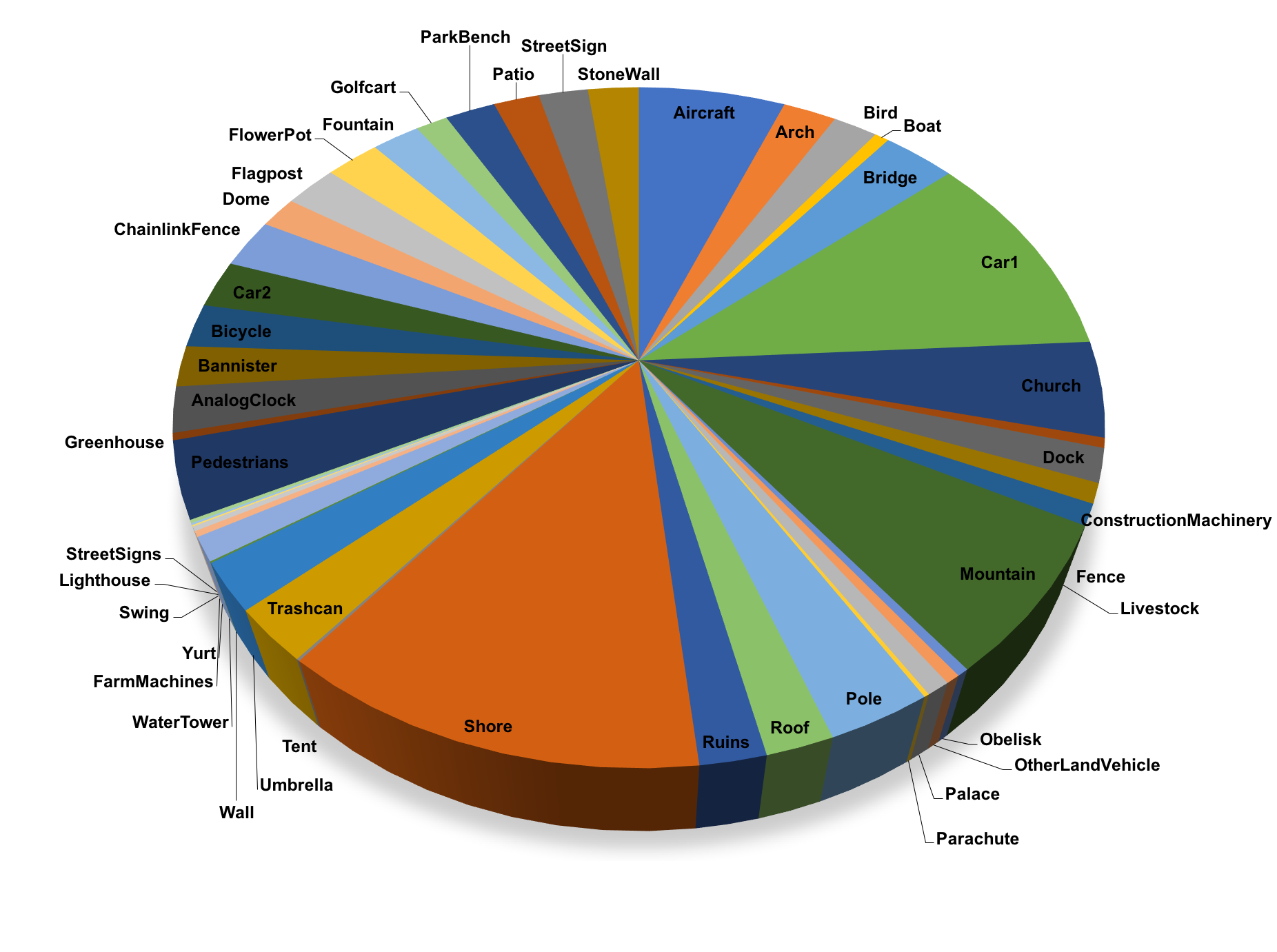}
    \caption[Shared Class Distribution for Object Detection Dataset]{Distribution of annotated images belonging to different classes in the UG$^2+$ Object Detection dataset. Our dataset is imbalanced due to some classes (\textit{e.g.}, ``Yurt," ``Wall," ``Tent") that have fewer than 100 images in the training dataset.}
    \label{fig:classdistribution_objdet}
\end{figure}

\textbf{Shared Class Distribution for the UG$^2 +$ Object Detection Dataset.}
Fig.~\ref{fig:classdistribution_objdet} shows the distribution of annotated images belonging to different classes in the Object Detection dataset. As can be seen from the figure, the dataset is imbalanced due to the aerial videos captured via gliders and UAVs, mostly in urban and rural areas. Hence we find an abundance of classes such as ``Mountains,'' ``Cars,'' and ``Aircraft'' in the dataset. Note that some classes, as for example, ``Yurt,'' ``Wall,'' and ``Tent'' have fewer than 100 images in the training dataset and therefore, are under-represented in the training dataset. We prioritized these classes in the test dataset by including more videos representing them in the dataset. As a result, the test dataset is far more challenging than the validation dataset. Also, class ``StreetSign'' in the Ground collection contains only stop sign images, and is kept separate from the images for that same class in the UAV and Glider collections, which contain additional signs. 

\begin{table}[t]
\begin{center}
\begin{tabular}{|c||c|c|c|c|}
    \hline
    \textbf{Dataset} & \textbf{Collection} & \textbf{UAV} & \textbf{Glider}  & \textbf{Ground}   \\
    \hline\hline
    \multirow{3}{*}{Training}& Frames & 26,105 & 19,152 & 41,227 \\
    & Extract. Objs. & 30,051 & 22,390 & 41,227 \\
    & UG$^2$ Classes & 31 & 14 & 19\\\hline
    \multirow{3}{*}{Testing} & Frames & 1,034 & 624 & 1,200 \\ 
    & Extract. Objs. & 1,034 & 624 & 1,200 \\
    & UG$^2$ Classes & 13 & 4 & 6  \\
    \hline
\end{tabular}\vspace{-4mm}
\end{center}
\caption{UG$^2$+ Object Detection Dataset Statistics.}\label{tab:detection_dataset}\vspace{-4mm}
\end{table}

\textbf{Sequestered Testing Procedure.}
The test set has a total of $2,858$ images and annotations from all three collections and is similar in all aspects with the training dataset (in terms of sensors used and object classes). The bottom row of Table~1 provides the testing dataset statistics. To make it more challenging for generalization, the classes were selected based on the difficulty of detecting them on the validation set (see Sec.~\ref{sec:baselines}, Fig.~1, and the description in Supp.~Sec.~1.1 for details related to this). For example, with respect to the Ground collection, the test dataset has objects whose distance from the camera was maximum or had the highest induced motion blur or toughest weather condition (\textit{e.g.}, rainy day, snowfall) or are underrepresented in the training dataset. We also prioritized classes which consistently have a low average precision in the validation set (see Supp.~Sec.~1.1). The evaluation for the formal challenge was sequestered, meaning participants did not have access to the test data prior to the submission of their algorithms. 
Each video in the test dataset was renamed to have a randomized filename to prevent participants from being able to use file metadata for classifying which collection the test frame came from.

\textbf{Evaluation Protocol for Detection.}
The objective of this challenge is to detect objects from a number of visual object classes in unconstrained environments. It is fundamentally a supervised learning problem in that a training set of labeled images is provided. Participants are not expected to develop novel object detection models. They are encouraged to use a pre-processing step (for instance, super-resolution, denoising, deblurring, or algorithms that jointly optimize image quality and recognition performance) in the detection pipeline. To evaluate the algorithms submitted by the participants, the raw frames of UG$^2$ are first pre-processed with the submitted algorithms and are then sent to the YOLOv3 detector~\citep{redmon2018yolov3}, which was fine-tuned on the UG$^2$ dataset. In this paper, we also consider another detector, YOLOv$2$~\citep{redmon2017yolo9000}. 

The metric used for scoring is Mean Average Precision (mAP) at Intersection over Union (IoU) $[0.15, 0.25, 0.5, 0.75, 0.9]$. The mAP evaluation is kept the same as PASCAL VOC~\citep{everingham2010pascal}, except for a single modification introduced in IoU. Unlike PASCAL VOC, we are evaluating mAP at different IoU values. This is to account for different sizes, scales of object and varying imaging artifacts in our dataset. We consider predictions to be ``a true match" when they share the same label and an $\mathrm{IoU} \geq  0.15, 0.25, 0.5, 0.75, 0.90$. The average precision (AP) for each class is calculated as the area under the precision-recall curve. Then, the mean of all AP scores is calculated, resulting in a mAP value from $0$ to $100\%$.

\vspace{-1mm}
\subsection{Object Classification Improvement in Video}\vspace{-0.5mm}

While interest in applying image enhancement techniques for classification purposes has started to grow, there has not been a direct application of such methods on video data. Currently, image enhancement algorithms attempt to estimate the visual aberrations $a$ of a given image $O$, in order to establish an aberration-free version $I$ of the scene captured (\textit{i.e.}, $O = I \otimes a + n$, where $n$ represents additional noise that might of be a byproduct of the visual aberration $a$). It is natural to assume that the availability of additional information --- like the information present in several contiguous video frames --- would enable a more accurate estimation of such aberrations, and as such a cleaner representation of the captured scene.

Taking this into account, we created challenge Track 1.2. The main goal of this track is to correct visual aberrations present in video in order to improve the classification results obtained with out-of-the-box classification algorithms. For this we adapted the evaluation method and metrics provided in~\citep{DBLP:journals/corr/abs-1901-09482} to take into account the temporal factor of the data present in the UG$^2$ dataset. Below we introduce the adapted training and testing datasets, as well as the evaluation metrics and baseline classification results for this task\footnote{The object classification dataset and evaluation kit are available from: http://bit.ly/UG2Devkit}.

\textbf{UG$^2$+ Classification Dataset.}
To leverage both the temporal and visual features of a given scene, we divided each of the $196$ videos of the original UG$^2$ dataset into multiple object sequences (for a total of $576$ object sequences). We define an object sequence as a collection of multiple frames in which a given object of interest is present in the camera view. For each of these sequences, we provide frame-level annotations detailing the location of the specified object (a bounding box with its coordinates) as well as the UG$^2$ class. 

\begin{table}[t]
\begin{center}
\begin{tabular}{|c||c|c|c|c|}
    \hline
    \textbf{Dataset} & \textbf{Collection} & \textbf{UAV} & \textbf{Glider}  & \textbf{Ground}   \\
    \hline\hline
    \multirow{3}{*}{Training}& Frames & 22,471 & 21,367 & 75,367 \\
    & Object Seqs. & 219 & 173 & 77 \\
    & UG$^2$ Classes & 29 & 19 & 20\\\hline
    \multirow{3}{*}{Validation}& Frames & 7,806 & 5,251 & 19,729 \\
    & Object Seqs. & 53 & 38 & 51 \\
    & UG$^2$ Classes & 20 & 15 & 20\\\hline
    \multirow{3}{*}{Testing} & Frames & 1,147 & 1,098 & 1,000 \\ 
    & Object Seqs. & 35 & 32 & 25 \\
    & UG$^2$ Classes & 14 & 7 & 20  \\
    \hline
\end{tabular}\vspace{-3mm}
\end{center}
\caption{UG$^2$+ Object Classification Dataset Statistics.}\label{tab:classification_dataset}
\end{table}

A UG$^2$ class encompasses a number of ImageNet classes belonging to a common hierarchy (\textit{e.g.}, the UG$^2$ class ``car" includes the ImageNet classes ``jeep," ``taxi," and ``limousine"), and is used in place of such classes to account for instances in which it might be impossible to identify the fine-grained ImageNet class that an object belongs to. For example, it might be impossible to tell what the specific type of car on the ground is from an aerial video where that car is hundreds --- if not thousands --- of feet away from the sensor.

Table~\ref{tab:classification_dataset} details the number of frames and object sequences extracted from each of the UG$^2$ collections for the training and testing datasets. An important difference between the training and testing datasets is that while some of the collections in the training set have a larger number of object sequences, that does not necessarily translate to a larger number of frames (as is the case with the UAV collection). As such, the number of frames (and thus duration) of each object sequence is not uniform across all three collections. The number of frames per object sequence can range anywhere from five frames to hundreds of frames. However, for the testing set, all of the object sequences have at least $40$ frames. It is important to note that while the testing set contains imagery similar to that present in the training set, the quality of the videos might vary. This results in differences in the classification performance (more details on this are discussed in Sec.~\ref{Sec:Classif:Baseline}).

\textbf{Evaluation Protocol for Video Classification.}
Given the nature of this sub-challenge, each pre-processing algorithm is provided with a given set of object sequences rather than individual --- and possibly unrelated --- frames. The algorithm is then expected to make use of both temporal and visual information pertaining to each object sequence in order to provide an enhanced version of each of the sequence's individual frames. The object of interest is then cropped out of the enhanced frames and used as input to an off-the-shelf classification network. For the challenge evaluation, we focused solely on VGG16 trained on ImageNet. However, we do provide a comparative analysis of the effect of the enhancement algorithms on different classifiers in Sec.~\ref{sec:results}. There was no fine-tuning on UG$^2$+ in the competition, as we were interested in making low-quality images achieve better classification results on a network trained with generally good quality data (as opposed to having to retrain a network to adapt to each possible artifact encountered in the real world). But we do look at the impact of fine-tuning in this paper.

The network provides us with a $1,000 \times n$ vector, where $n$ corresponds to the number of frames in the object sequence, detailing the confidence score of each of the $1,000$ ImageNet classes on each of the sampled frames. To evaluate the classification accuracy of each object sequence we use Label Rank Average Precision (LRAP) \citep{tsoumakas2009mining}:
\[\mathrm{LRAP}(y,\hat{f})=\frac{1}{n}\sum_{i=0}^{n}\frac{1}{\left | y_{i} \right |} \sum_{j:y_{k}=1}\frac{\left | L_{ij} \right |}{rank_{ij}} \]

\[L_{ij} = \left \{ k: y_{ik} = 1, \hat{f_{ik}} > \hat{f_{ij}} \right \}\]

\[rank_{ij} = \left | \left \{ k : \hat{f_{ik}} \geq \hat{f_{ij}} \right \} \right | \]

LRAP measures the fraction of highly ranked ImageNet labels (\textit{i.e.}, labels with the highest confidence score $\hat{f}$ assigned by a given classification network, such as VGG16) $L_{ij}$ that belong to the true label UG$^2$ class $y_{i}$ of a given sequence $i$ containing $n$ frames. A perfect score (LRAP $= 1$) would then mean that all of the highly ranked labels belong to the ground-truth UG$^2$ class. For example, if the class ``shore" has two sub-classes lakeshore and seashore, then the top $2$ predictions of the network for all of the cropped frames in the object sequence are in fact lakeshore and seashore.
LRAP is generally used for multi-class classification tasks where a single object might belong to multiple classes. Given that our object annotations are not as fine-grained as the ImageNet classes (each of the UG$^2$ classes encompasses several ImageNet classes), we found this metric to be a good fit for our classification task. 
\section{Challenge Workshop Entries}\vspace{-2mm}

Here we describe the approaches for one or both of the evaluation tasks from each of the 2019 UG$^2+$ challenge participants.

\textbf{IMECAS-UMAC: Intelligent Resolution Restoration.}
The main objective of the algorithms submitted by team IMECAS-UMAC was to restore resolution based on scene content with deep learning. As UG$^2$ contains videos with varying degrees of imaging artifacts coming from three different collections, their method incorporated a scene classifier to predict which collection the image came from in order to apply targeted enhancement and restoration to that image. For the UAV and Ground collection respectively, they applied the VDSR~\cite{Kim:2016:VDSR} super-resolution algorithm and Fast Artifact Reduction CNN~\cite{FastARCNN}. For the Glider collection, they chose to do nothing.

\textbf{UCAS-NEU: Smart HDR.} Team UCAS-NEU concentrated on enhancing the resolution and dynamic range of the videos in UG$^2$ via deep learning. Irrespective of the collection the images came from, they used linear blending to add the image with its corresponding Gaussian-blurred counterpart to perform a temporal cross-dissolve between these two images, resulting in a sharpened output. Their other algorithm used the Fast Artifact Reduction CNN~\cite{FastARCNN} to reduce compression artifacts present in the UG$^2$ dataset, especially in the UAV collection caused by repeated uploads and downloads from YouTube.

\textbf{Honeywell: Camera and Conditions-Relevant Enhancements (CCRE).}
Team Honeywell used their CCRE algorithm~\cite{vidalmata2019bridging} to closely target image enhancements to avoid the counter-productive results that the UG$^2$ dataset has highlighted~\cite{vidal2018ug}. Their algorithm relies on the fact that not all types of enhancement techniques {\em may} be useful for a particular image coming from any of the three different collections of the UG$^2$ dataset. To find the useful subset of image enhancement techniques required for a particular image, the CCRE pipeline considers the intersection of camera-relevant enhancements with conditions-relevant enhancements. Examples of camera-relevant enhancements include de-interlacing, rolling shutter removal (both depending on the sensor hardware), and de-vignetting (for fisheye lenses). Example conditions-relevant enhancements include de-hazing (when imaging distant objects outdoors) and raindrop removal. While interlacing was the largest problem with the Glider images, the Ground and UAV collections were degraded by compression artifacts. De-interlacing was performed on detected images from the Glider dataset with the expectation that the edge-type features learned by the VGG network will be impacted by jagged edges from interlacing artifacts. Detected video frames from the UAV and Ground collections were processed with the Fast Artifact Reduction CNN~\cite{FastARCNN}.

For their other algorithms, they used an autoencoder trained on the UG$^2$ dataset to enhance images, and a combination of autoencoder and de-interlacing algorithm to enhance de-interlaced images. The encoder part of the autoencoder follows the architecture of SRCNN~\cite{dong2014learning} 

\textbf{TUM-CAS: Adaptive Enhancement.}
Team TUM-CAS employed a method similar to IMECAS-UCAS. They used a scene classifier to predict which collection the image came from, or reverted to a ``None" failure case. Based on the collection, they used a de-interlacing technique similar to Honeywell's for the Glider collection, an image sharpening method and histogram equalization for the Ground collection, and histogram equalization followed by super-resolution~\cite{Kim:2016:VDSR} for the UAV collection. If the image was found to not belong to any of the collections in UG$^2$, they chose to do nothing.

\textbf{Meitu's MTLab: Data Driven Enhancement.}
Meitu's MTLab proposed an end-to-end neural network incorporating direct supervision through the use of traditional cross-entropy loss in addition to the YOLOv$3$ detection loss in order to improve the detection performance. The motivation behind doing this is to make the YOLO detection performance as high as possible, \textit{i.e.}, enhance the features of the image required for detection. The proposed network consists of two sub-networks: Base Net and Combine Net. For Base Net, they used the convolutional layers of ResNet~\cite{he2016deep} as their backbone network to extract the features at different levels from the input image. Features from each convolution branch are then passed to the individual layers of Combine Net, which are fused to get an intermediate image. The final output is created as an addition of the intermediate image from Combine Net and the original image. While they use ResNet as the Base Net for its strong feature extraction capability, the Combine Net captures the context at different levels (low-level features like edges to high-level image statistics like object definitions). For training this network end-to-end, they used cross-entropy and YOLOv3 detection loss and UG$^2$ as the dataset.

\textbf{Sunway.AI: Sharpen and Histogram Equalization with Super-resolution (SHESR).}
Team Sunway.AI employed a method similar to IMECAS-UCAS and TUM-CAS. They also used a scene classifier to predict which collection the image came from, or reverted to a ``None" failure case. They used histogram equalization followed by super-resolution~\cite{Kim:2016:VDSR} for the UAV collection, and image sharpening followed by Fast Artifact Reduction CNN~\cite{FastARCNN} for the Ground collection to remove blocking artifacts due to JPEG compression. If the image was from the Glider collection or was found to not belong to any of the collections in UG$^2$, they chose to do nothing.

\begin{table}[t]
\setlength\tabcolsep{4pt}
\begin{tabular}{|r|l|l|l|l|l|l|}
\hline

\multicolumn{1}{|l|}{} & \multicolumn{2}{c|}{\textbf{UAV}} & \multicolumn{2}{c|}{\textbf{Glider}} & \multicolumn{2}{c|}{\textbf{Ground}} \\ \cline{2-7} 
\multicolumn{1}{|l|}{\textbf{mAP}} & \multicolumn{1}{c|}{\textbf{Val.}} & \multicolumn{1}{c|}{\textbf{Test}} & \multicolumn{1}{c|}{\textbf{Val.}} & \multicolumn{1}{c|}{\textbf{Test}} & \multicolumn{1}{c|}{\textbf{Val.}} & \multicolumn{1}{c|}{\textbf{Test}} \\ \hline

\textbf{@15} & 96.4\%   & 1.3\%     & 95.1\%   & 5.19\%    & 100\%     & 31.6\% \\ \hline

\textbf{@25} & 95.5\%   & 1.3\%     & 94.8\%   & 5.19\%    & 100\%     & 31.6\% \\ \hline

\textbf{@50} & 88.6\%   & 0.61\%    & 91.1\%   & 0.01\%    & 100\%     & 21.5\% \\ \hline

\textbf{@75} & 39.5\%   & 0\%       & 40.3\%   & 0\%       & 96.7\%   & 15.8\% \\ \hline

\textbf{@90} & 1.9\%    & 0\%       & 2.9\%    & 0\%       & 54.7\%   & 0.04\% \\ \hline

\end{tabular}
\caption{mAP scores for the UG$^2$ Object Detection dataset with YOLOv$3$ fine-tuned on the UG$^2$ dataset. For the mAP scores for YOLOv2, see Table~2 in the Supp. Mat.}
\label{tab:dataset_mAP}
\vspace{-6mm}
\end{table}

\vspace{-1mm}
\section{Results \& Analysis}\vspace{-2mm}
\label{sec:results}
In  the  following  section,  we  review  the  results  that  came out of the UG$^2$+ Challenge held at CVPR 2019, and discuss additional results from the slate of baseline algorithms. The challenge received $16$ enhancement algorithms, developed by the six teams to address the detection and classification tasks defined in Sec.~\ref{Challenge}.  
\vspace{-1mm}
\subsection{Baseline Results}\vspace{-0.5mm}\label{sec:baselines}

\textbf{Object Detection Improvement on Video.} 
In order to establish scores for detection performance before and after the application of image enhancement and restoration algorithms submitted by participants, we use the YOLOv$3$ object detection model~\citep{redmon2018yolov3} to localize and identify objects in a frame and then consider the mAP scores at IoU $[0.15, 0.25, 0.5,0.75,0.9]$. Since the primary goal of our challenge does not involve developing a novel detection method or comparing the performance among popular object detectors, ideally, any detector could be used for measuring the performance. We chose YOLOv$3$ for the challenge because it is easy to train and is the fastest among the popular off-the-shelf detectors~\citep{girshick2015fast,liu2016ssd,lin2017focal}. 

We fine-tuned YOLOv$3$ to reflect the UG$^2$+ object detection classes and measured its performance on the validation and test data per collection to establish baseline performance.
Table~\ref{tab:dataset_mAP} shows the baseline mAP scores obtained using YOLOv$3$ on raw video frames (\textit{i.e.}, without any pre-processing). Overall, we observe distinct differences between the results for all three collections, particularly between the airborne collections (UAV and Glider) and the Ground collection. Since the network was fine-tuned with UG$^2+$, we expected the mAP score at $0.5$ IoU for validation to be fairly high for all three collections. The Ground collection receives a perfect score of $100\%$ for mAP at $0.5$. This is due to the fact that images within the Ground collection have minimal imaging artifacts and variety, as well as many pixels on target, compared to the other collections. The UAV collection, on the other hand, has the worst performance due to relatively small object scales and sizes, as well as compression artifacts resulting from the processing applied by YouTube. It achieves a very low score of $1.93\%$ for mAP at $0.9$.

For the test dataset, we concentrated more on the classes of UG$^2$ that were underrepresented in the training dataset to make it more challenging based on the distribution shown in Fig.~\ref{fig:classdistribution_objdet}. We also intentionally selected difficult conditions. For example, for the Ground dataset, we concentrated on objects whose distance from the camera was maximum ($200$ ft) or had the highest induced motion blur ($180$ rpm) or toughest weather condition (\textit{e.g.}, rainy day, snowfall). Correspondingly, the mAP scores on the test dataset are very low. At operating points of $0.75$ and $0.90$ IoU, most of the object classes in the Glider and UAV collections are unrecognizable. This, however, varies for the Ground collection, which receives a score of $15.75\%$ and $0.04\%$ respectively. The classes that were readily identified in the ground collection were large objects: ``Analog Clock," ``Arch," ``Street Sign," and ``Car 2" for $0.75$ IoU and only ``Arch" for $0.90$ IoU. We also fine-tuned a separate detector, YOLOv$2$~\citep{redmon2017yolo9000}, on UG$^2+$ to assess the impact of a different detector architecture on our dataset. The details can be found in Supp.~Sec.~1.2.

\begin{figure*}[ht]
\centering
    \begin{subfigure}[]{0.32\textwidth}
        \begin{scaletikzpicturetowidth}{\textwidth}
        \begin{tikzpicture}[scale=\tikzscale]
            \begin{axis}[
                nodes near coords,
                ybar,
                every node near coord/.append style={rotate=90, anchor=west, font=\small},
                enlarge y limits={upper,value=1.5},
                legend style={at={(0.5,1.15)},
                  anchor=north,legend columns=-1},
                ylabel={mAP (\%)},
                y label style={at={(0.05,0.5)}},
                symbolic x coords={UCAS-NEU, IMECAS-UMAC, Honeywell Intl, TUM-CAS, Sunway.AI, MTlab, Baseline},
                xtick=data,
                bar width = 6pt,
                xticklabel style={anchor= east, align = right, rotate=25, font=\small},
                ]
                \addplot [black!100!orange, fill=orange!100] coordinates {(UCAS-NEU, 1.32) (IMECAS-UMAC, 1.12) (Honeywell Intl, 1.30) (TUM-CAS, 1.26) (Sunway.AI, 1.24) (MTlab, 0.59) (Baseline, 1.3)};
                \addplot [black!60!green,fill=black!40!green] coordinates {(UCAS-NEU, 1.32) (IMECAS-UMAC, 1.12) (Honeywell Intl, 1.30) (TUM-CAS, 1.26) (Sunway.AI, 1.24) (MTlab, 0.59) (Baseline, 1.3)};
                \legend{mAP@0.15, mAP@0.25}
            \end{axis}
        \end{tikzpicture}
        \end{scaletikzpicturetowidth}
        \caption{UAV collection} \label{fig:uav_det_top_results}
    \end{subfigure} 
    \begin{subfigure}[]{0.32\textwidth}
        \begin{scaletikzpicturetowidth}{\textwidth}
        \begin{tikzpicture}[scale=\tikzscale]
            \begin{axis}[
                nodes near coords,
                ybar,
                every node near coord/.append style={rotate=90, anchor=west, font=\small},
                enlarge y limits={upper,value=1.5},
                legend style={at={(0.5,1.15)},
                  anchor=north,legend columns=-1},
                ylabel={mAP (\%)},
                y label style={at={(0.05,0.5)}},
                symbolic x coords={UCAS-NEU, IMECAS-UMAC, Honeywell Intl, TUM-CAS, Sunway.AI, MTlab, Baseline},
                xtick=data,
                bar width = 6pt,
                xticklabel style={anchor= east, align = right, rotate=25, font=\small }
                ]
                \addplot [black!60!green,fill=black!40!green] coordinates {(UCAS-NEU, 5.19) (IMECAS-UMAC, 5.19) (Honeywell Intl, 5.19) (TUM-CAS, 5.19) (Sunway.AI, 5.19) (MTlab, 2.57) (Baseline, 5.19)};
                \addplot [black!90!blue, fill=blue!90] coordinates {(UCAS-NEU, 0.02) (IMECAS-UMAC, 0.01) (Honeywell Intl, 0.01) (TUM-CAS, 0.01) (Sunway.AI, 0.01) (MTlab, 1.02) (Baseline, 0.01)};
                \legend{mAP@0.25, mAP@0.5}
            \end{axis}
        \end{tikzpicture}
        \end{scaletikzpicturetowidth}
        \caption{Glider collection} \label{fig:gli_det_top_results}
    \end{subfigure} 
    \begin{subfigure}[]{0.32\textwidth}
        \begin{scaletikzpicturetowidth}{\textwidth}
        \begin{tikzpicture}[scale=\tikzscale]
            \begin{axis}[
                nodes near coords,
                ybar,
                every node near coord/.append style={rotate=90, anchor=west, font=\small},
                enlarge y limits={upper,value=1.5},
                legend style={at={(0.5,1.15)},
                  anchor=north,legend columns=-1},
                ylabel={mAP (\%)},
                y label style={at={(0.05,0.5)}},
                symbolic x coords={UCAS-NEU, IMECAS-UMAC, Honeywell Intl, TUM-CAS, Sunway.AI, MTlab, Baseline},
                xtick=data,
                bar width = 6pt,
                xticklabel style={anchor= east, align = right, rotate=25, font=\small }
                ]
            \addplot [black!60!cyan,fill=black!40!cyan] coordinates {(UCAS-NEU, 16.07) (IMECAS-UMAC, 15.81) (Honeywell Intl, 15.75) (TUM-CAS, 15.81) (Sunway.AI, 15.81) (MTlab, 13.08) (Baseline, 15.75)};
            \legend{mAP@0.75}    
            \end{axis}
        \end{tikzpicture}
        \end{scaletikzpicturetowidth}
        \caption{Ground collection @ mAP 0.75. } \label{fig:grnd_det_top_results_75}
    \end{subfigure} 
    \caption{Object Detection: Best performing submissions per team across three different mAP intervals.}
    \vspace{-3mm}
\end{figure*}
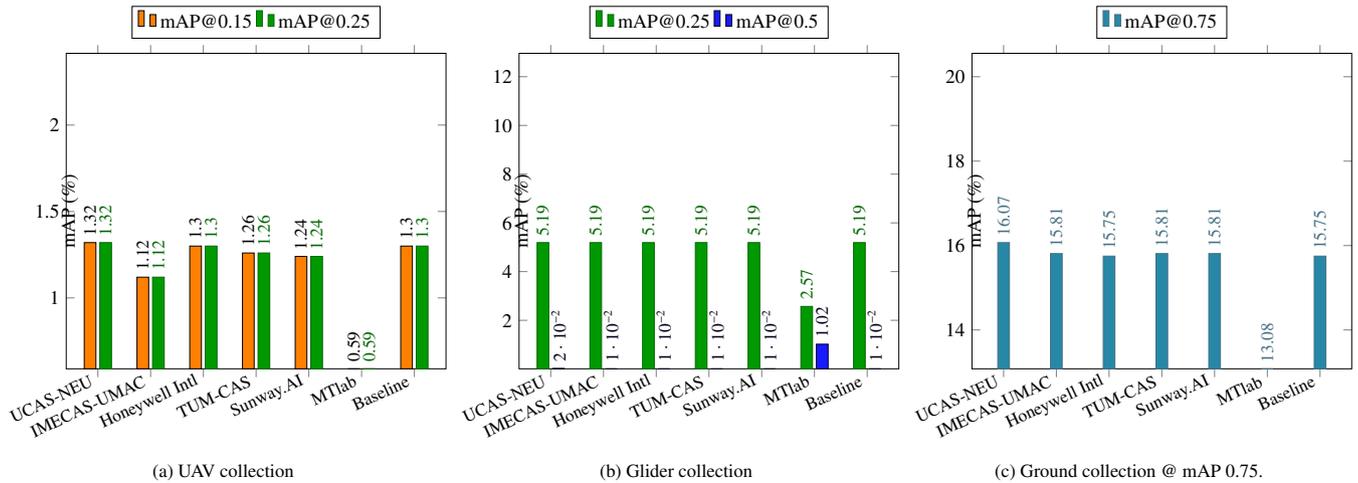

\textbf{Object Classification Improvement on Video.} \label{Sec:Classif:Baseline}
Table~\ref{tab:classification_baselines} shows the average LRAP of each of the collections on both the training and testing datasets without any restoration or enhancement algorithm applied to them. These scores were calculated by averaging the LRAP score of each of the object sequences $y_{c}$ of a given UG$^2$ class $C_i$, for all the $k$ classes in that particular collection $D$:\vspace{-2mm}
\[ \mathrm{AverageLRAP}(D) = \frac{1}{K}\sum_{i=0}^{K} \mathrm{LRAP}(C_{i}) \]
\[ \mathrm{LRAP} (C_{i}) = \frac{1}{|C_i|}\sum_{c=0}^{|C_i|} \mathrm{LRAP} (y_{c}, \hat{f})  \mid C_i \in D_{classes}\]

As can be observed from the training set, the average LRAP scores for each collection tend to be quite low, which is not surprising given the challenging nature of the dataset. While the Ground dataset presents a higher average LRAP, the scores from the two aerial collections are very low. This can be attributed to both aerial collections containing more severe artifacts as well as a vastly different capture viewpoint than the one in the Ground collection (whose images would have a higher resemblance to the classification network training data). It is important to note the sharp difference in the performance of different classifiers on our dataset. While the ResNet50~\citep{DBLP:journals/corr/HeZRS15} classifier obtained a slightly better but similar performance to the VGG16 classifier --- which was the one used to evaluate the performance of the participants in the challenge --- other networks such as DenseNet~\citep{DBLP:journals/corr/HuangLW16a}, MobileNet~\citep{DBLP:journals/corr/HowardZCKWWAA17}, and NASNetMobile~\citep{DBLP:journals/corr/ZophVSL17} perform poorly when classifying our data. It is likely that these models are highly oriented to ImageNet-like images, and have more trouble generalizing to our data without further fine-tuning.

\begin{table}[t]
\setlength\tabcolsep{4pt}
\begin{tabular}{|r|l|l|l|l|l|l|}
\hline

\multicolumn{1}{|l|}{} & \multicolumn{2}{c|}{\textbf{UAV}} & \multicolumn{2}{c|}{\textbf{Glider}} & \multicolumn{2}{c|}{\textbf{Ground}} \\ \cline{2-7} 
\multicolumn{1}{|l|}{} & \multicolumn{1}{c|}{\textbf{Train}} & \multicolumn{1}{c|}{\textbf{Test}} & \multicolumn{1}{c|}{\textbf{Train}} & \multicolumn{1}{c|}{\textbf{Test}} & \multicolumn{1}{c|}{\textbf{Train}} & \multicolumn{1}{c|}{\textbf{Test}} \\ \hline

V16 & 12.2\% & 12.7\% & 10.7\% & 33.7\% & 46.3\% & 29.4\% \\ \hline

R50 & 13.3\% & 15.1\% & 11.7\% & 28.5\% & 51.7\% & 38.7\% \\ \hline

D201 & 3.9\% & 2.1\% & 3.9\% & 1.0\% & 7.5\% & 3.8\% \\ \hline

MV2 & 1.8\% & 1.8\% & 1.5\%& 1.2\% & 6.9\% & 5.4\% \\ \hline

NNM & 1.2\% & 2.0\% & 1.2\%& 0.5\% & 1.0\% & 0.5\% \\ \hline

\end{tabular}
\caption{UG$^2$ Object Classification Baseline Statistics for ImageNet pre-trained networks: VGG16 (V16), ResNet50 (R50), DenseNet201 (D201), MobileNetV2 (MV2), NASNetMobile (NNM).}
\label{tab:classification_baselines}
\vspace{-5mm}
\end{table}

For the testing set, the UAV collection maintains a low score. However, the Ground collection's score drops significantly. This is mainly due to a higher amount of frames with problematic conditions (such as rain, snow, motion blur or just an increased distance to the target objects), compared to the frames in the training set. A similar effect is shown on the Glider collection, for which the majority of the videos in the testing set tended to portray either larger objects (\textit{e.g.}, mountains) or objects closer to the camera view (\textit{e.g.}, other aircraft flying close to the video-recording glider). When exclusively comparing the classification performance of the common classes of the two datasets, we observe a significant improvement in the average LRAP of the training set (with an average LRAP of $28.46$\% for the VGG16 classifier). More details on this analysis can be found in the Supp. Mat. Table 8.

\begin{figure}[h]
\center
\begin{scaletikzpicturetowidth}{0.32\textwidth}
\begin{tikzpicture}[scale=\tikzscale]
\begin{axis}[
    nodes near coords,
    ybar,
    every node near coord/.append style={rotate=90, anchor=west, font=\small},
    enlarge y limits={upper,value=0.3},
    legend style={at={(0.5,1.15)},
      anchor=north,legend columns=-1},
    ylabel={mAP (\%)},
    symbolic x coords={UCAS-NEU, IMECAS-UMAC, Honeywell Intl, TUM-CAS, Sunway.AI, MTlab, Baseline},
    xtick=data,
    bar width = 8pt,
    xticklabel style={anchor= east, align = right, rotate=25, font=\small }
    ]
\addplot [black!80!magenta, fill=magenta!80] coordinates {(UCAS-NEU, 0.04) (IMECAS-UMAC, 0.04) (Honeywell Intl, 0.01) (TUM-CAS, 0.04) (Sunway.AI, 0.04) (MTlab, 0.19) (Baseline, 0.01)};
\legend{mAP@0.90}
\end{axis}
\end{tikzpicture}
\end{scaletikzpicturetowidth}
\caption{Object Detection: Best performing submissions per team for the Ground collection @ mAP 0.9. 
}
\label{fig:grnd_det_top_results_9}
\vspace{-6mm}
\end{figure}
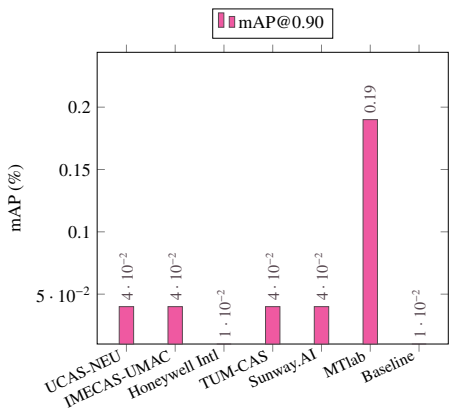

To evaluate the impact of the domain transfer between ImageNet features and our dataset, specifically looking at the disparity between the training and testing performance on the Glider collection, we fine-tuned the VGG16 network on the Glider collection training set for 200 epochs with a training/validation split of $80/20$\%, obtaining a training, validation, and testing accuracy of $91.67$\%, $27.55$\%, and $20.25$\% respectively. Once the network was able to gather more information about the dataset, the gap between validation and testing was diminished. Nevertheless, the broad difference between the training and testing scores indicates that the network has problems generalizing to the UG$^2$+ data, perhaps due to the large number of image aberrations present in each image.

\vspace{-1mm}
\subsection{Challenge Participant Results}\vspace{-0.5mm}

\textbf{Object Detection Improvement in Video.} For the detection task, each participant's algorithms were evaluated on the mAP score at IoU [$0.15, 0.25, 0.5, 0.75, 0.9$]. If an algorithm had the highest score or the second-highest score (in situations where the baseline had the best performance), in any of these metrics, it was given a score of $1$. The best performing team was selected based on the scores obtained by their algorithms. As in the 2018 challenge, each team was allowed to submit three algorithms. Thus, the upper bound for the best performing team is $45$: $3$ (algorithms) $\times$ $5$ (mAP at IoU intervals) $\times$ $3$ (collections).

Figs.~\ref{fig:uav_det_top_results}, \ref{fig:gli_det_top_results}, \ref{fig:grnd_det_top_results_75} and \ref{fig:grnd_det_top_results_9} show the results from the detection challenge for the best performing algorithms submitted by the participants for the different collections, as compared to the baseline. For brevity, only the results of the top-performing algorithm for each participant are shown. Full results can be found in Supp. Fig.~2. We also provide the participants' results on YOLOv$2$ in Supp.~Table~2.

We found the mAP scores at $[0.15, 0.25]$, $[0.25, 0.5]$, and $[0.75, 0.9]$ to be most discriminative in determining the winners. This is primarily due to the fact that objects in the airborne data collections (UAV, Glider) have small sizes and different degrees of views compared to the objects in the Ground data collection. In most cases, none of the algorithms could exceed the performance of the baseline by a considerable margin, re-emphasizing that detection in unconstrained environments is still an unsolved problem. 

The UCAS-NEU team with their strategy of sharpening images won the challenge by having the highest mAP scores for UAV (0.02\% improvement over baseline) and Ground dataset (0.32\% improvement). This shows that image sharpening can re-define object edges and boundaries, thereby helping in object detection. For the Glider collection, most participants chose to do nothing. Thus the results of several algorithms are an exact replica of the baseline result. Also, most participants tended to use a scene classifier trained on UG$^2$ to identify which collection the input image came from. The successful execution of such a classifier determines the processing to be applied to the images before sending them to a detector. If the classifier failed to detect the collection, no pre-processing would be applied to the image, sending the raw unchanged frame to the detector. Large numbers of failed detections likely contributed to some results being almost the same as the baseline.

\begin{figure}[t]
     \centering
     \begin{subfigure}[b]{0.5\textwidth}
         \includegraphics[width=0.3\textwidth]{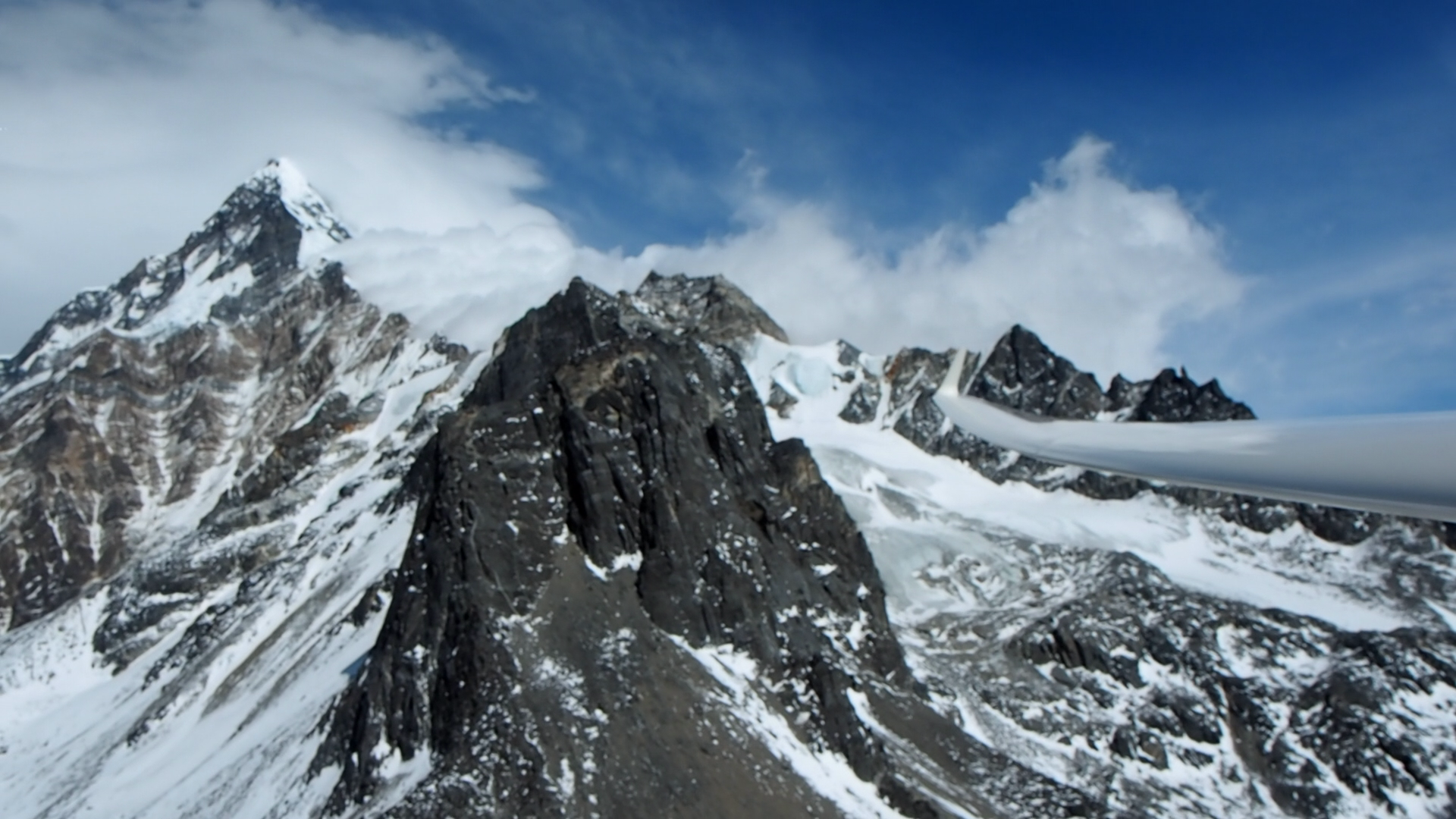}
         \includegraphics[width=0.3\textwidth]{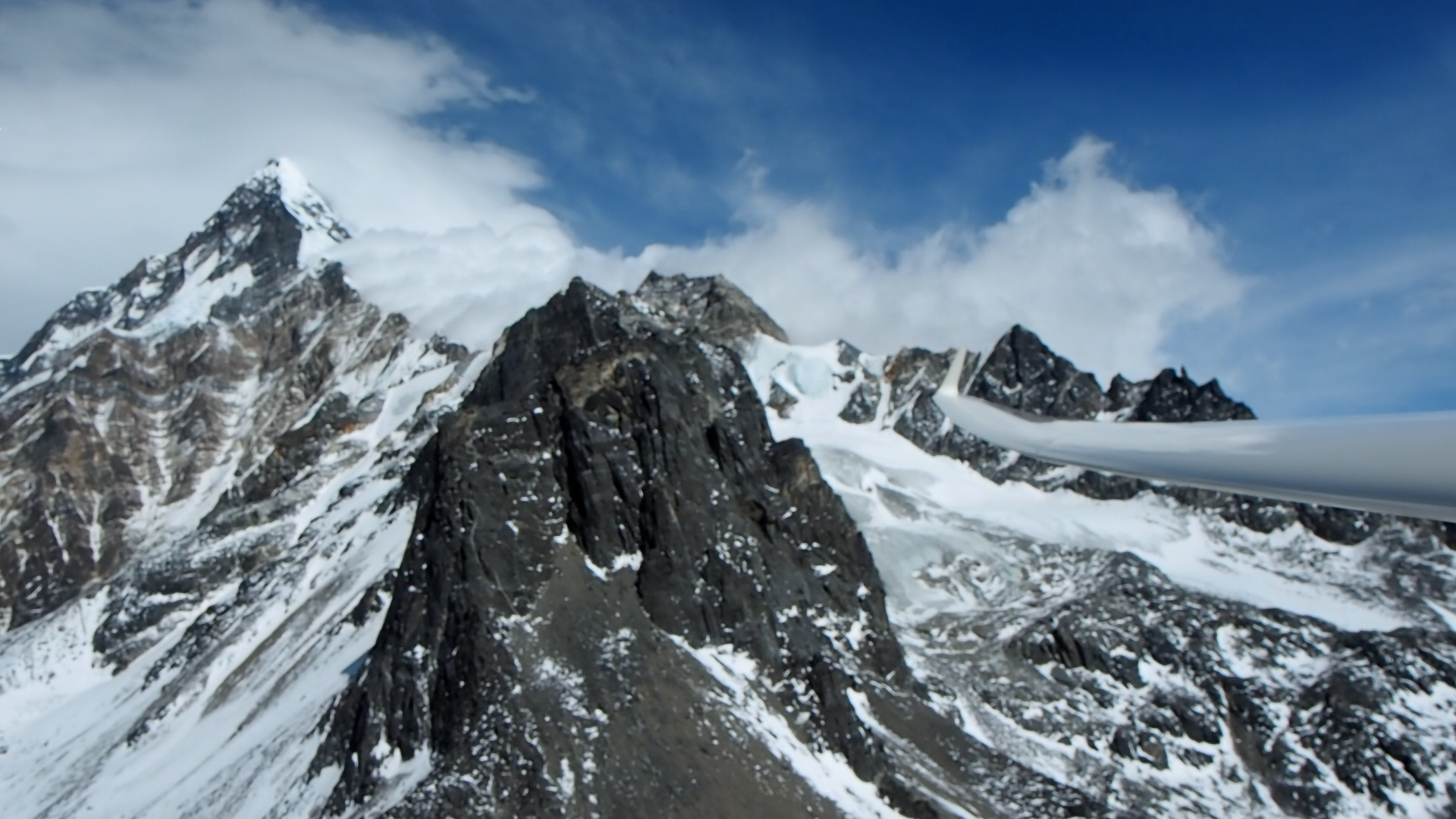}
         \includegraphics[width=0.3\textwidth]{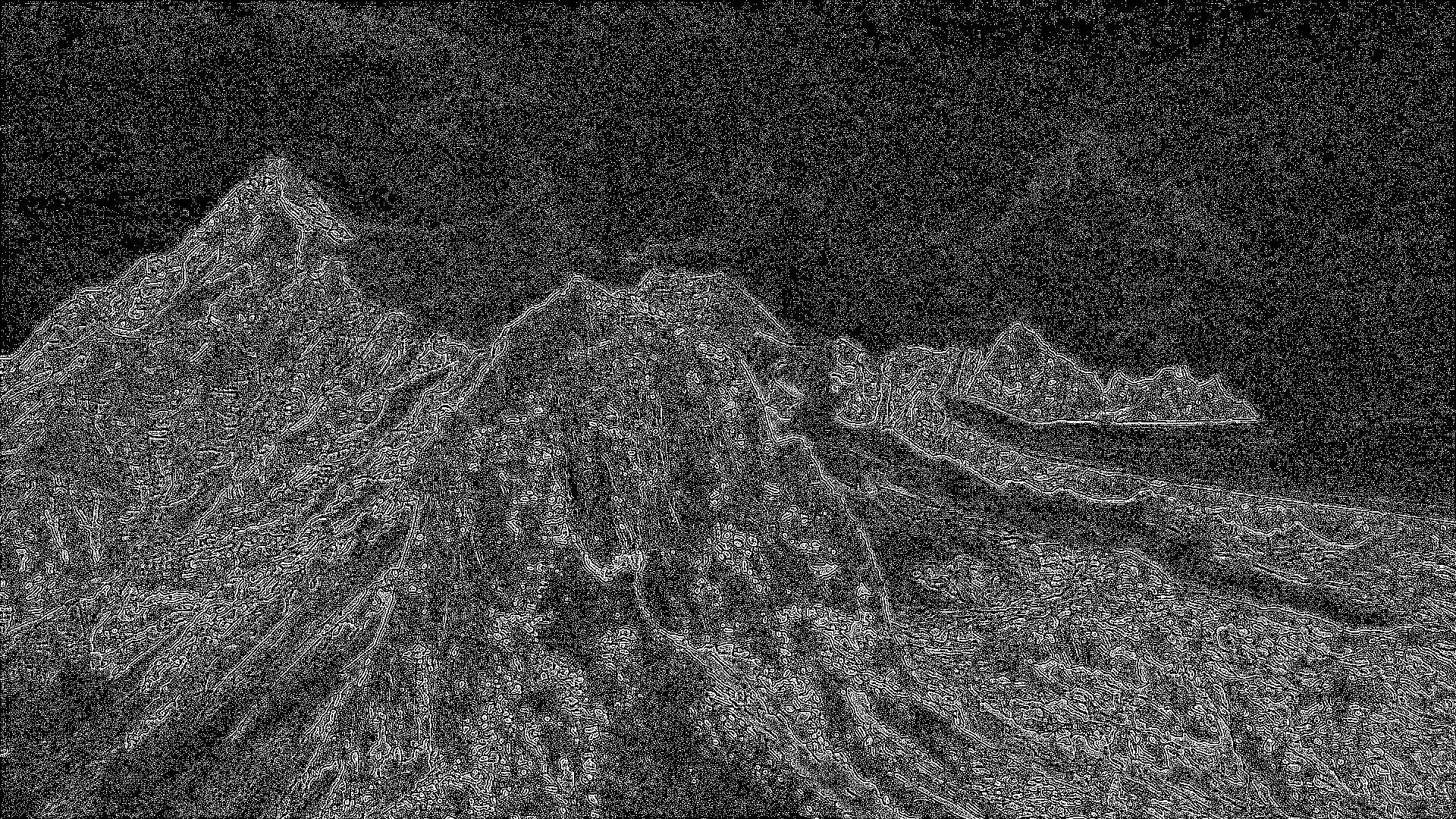}\caption{UCAS-NEU}
     \end{subfigure}
     \begin{subfigure}[b]{0.5\textwidth}
         \includegraphics[width=0.3\textwidth]{figures/o27493.png}
         \includegraphics[width=0.3\textwidth]{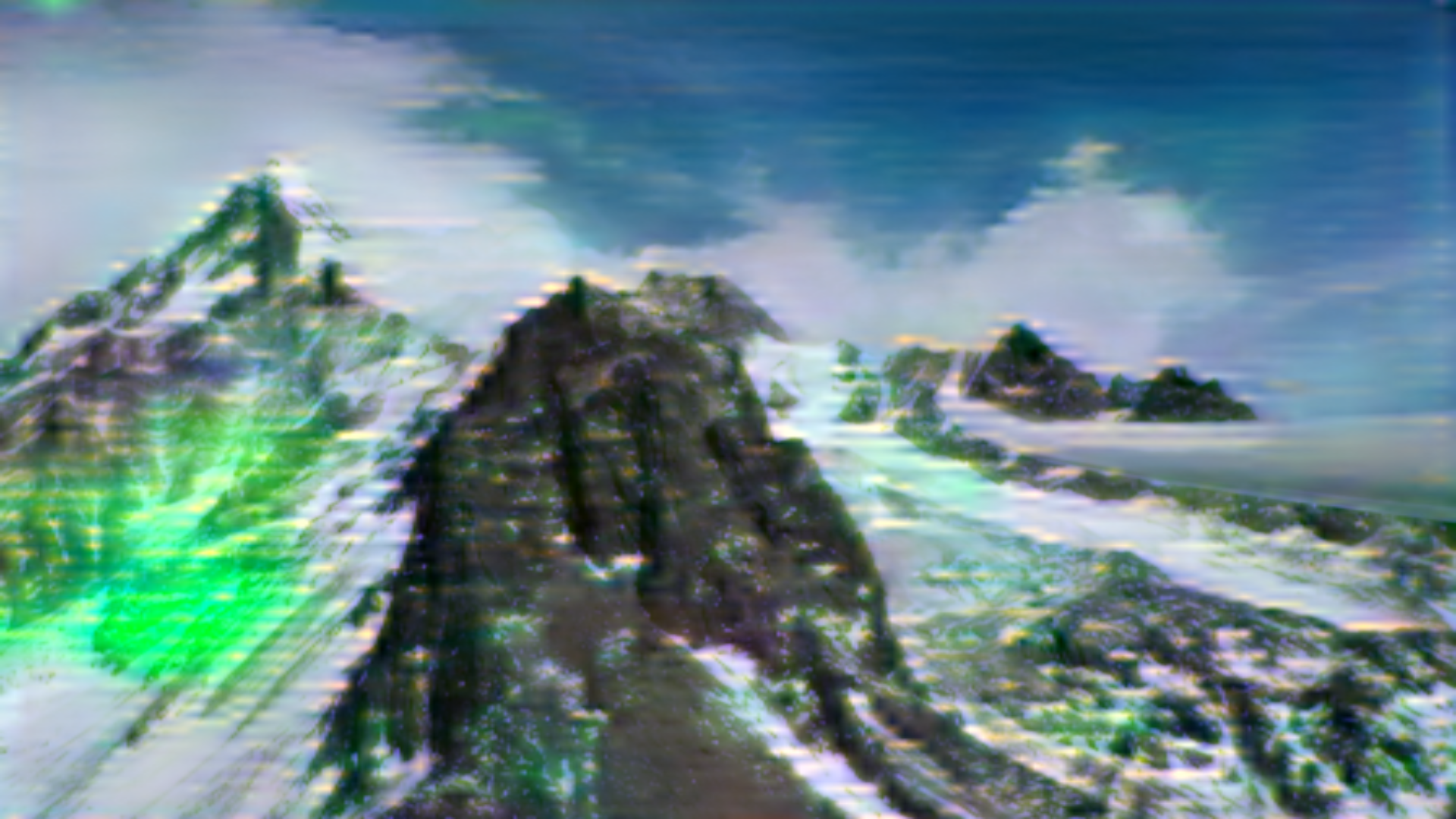}
         \includegraphics[width=0.3\textwidth]{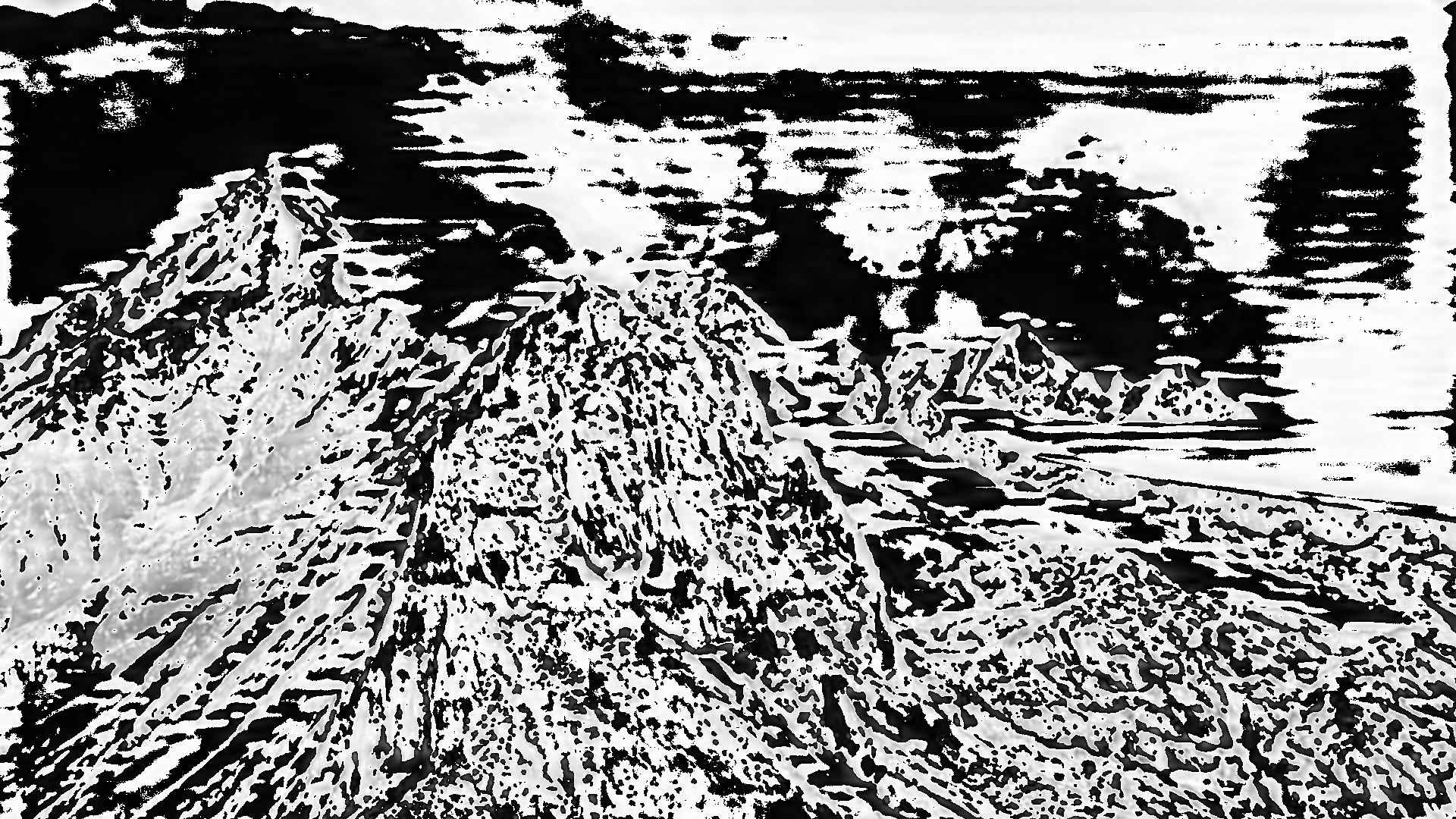}\caption{MTLab}
     \end{subfigure}
     \begin{subfigure}[b]{0.5\textwidth}
         \includegraphics[width=0.3\textwidth]{figures/o27493.png}
         \includegraphics[width=0.3\textwidth]{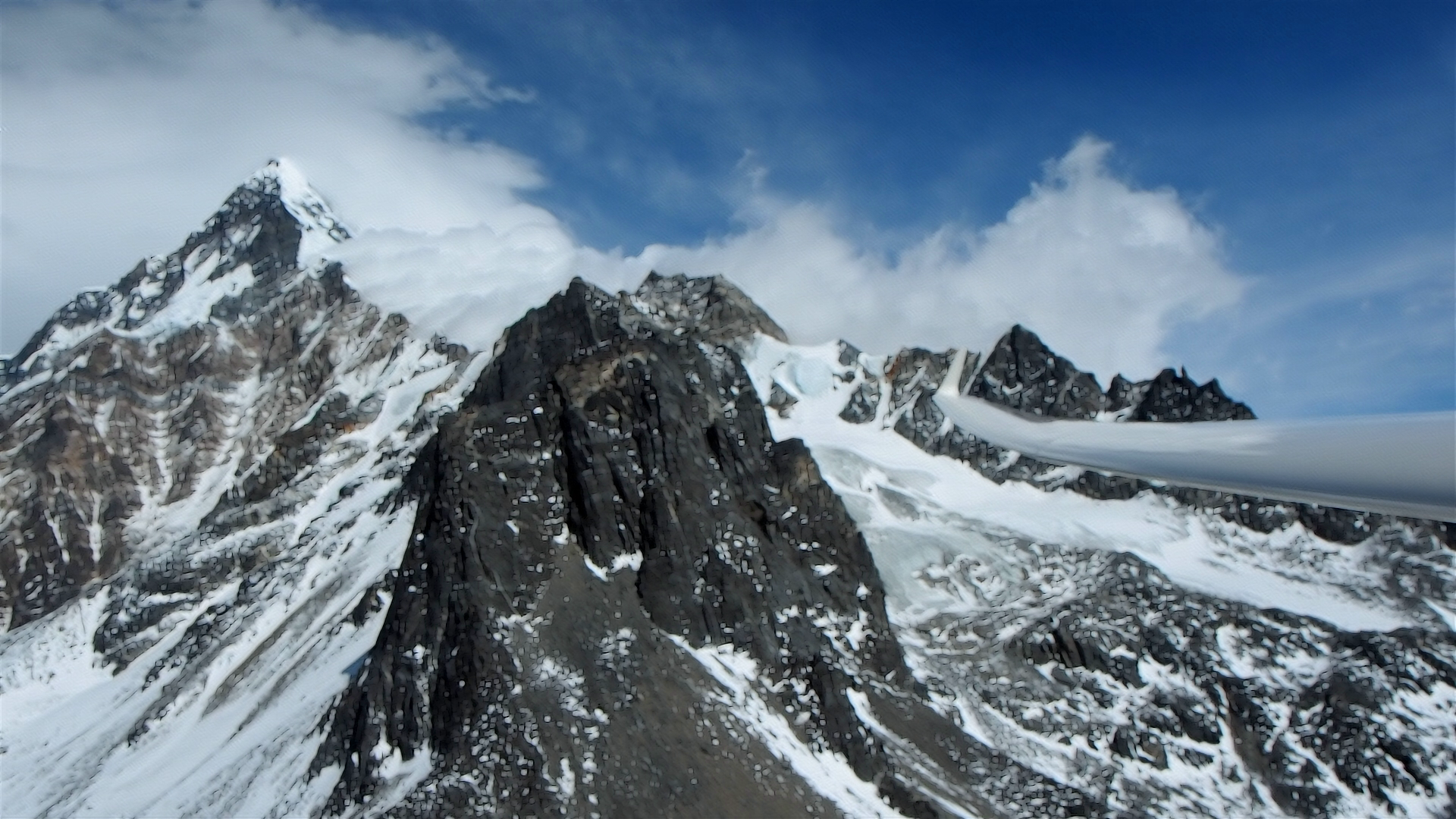}
         \includegraphics[width=0.3\textwidth]{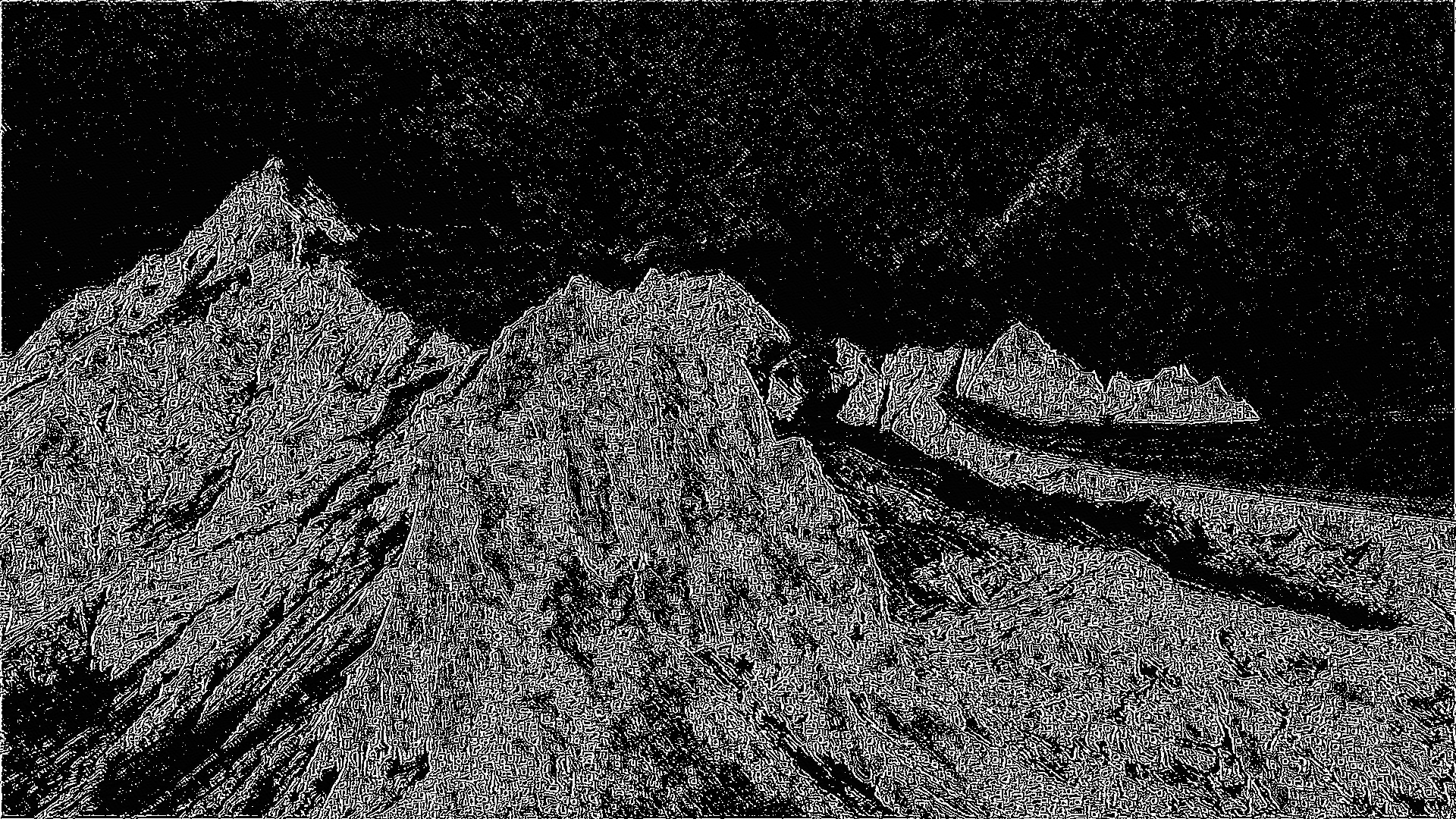}\caption{IMECAS-UMAC}
     \end{subfigure}\vspace{-2mm}
    \caption{Visual comparison of enhancement and restoration algorithms submitted by participants. The first image is the original input, second the algorithm's output and the third is the difference between the input and output. }
    \label{fig:alg_comp}    
    \vspace{-5mm}    
\end{figure}

An interesting observation here concerns the performance of the algorithm from MTLab. As discussed previously, MTlab's algorithm jointly optimized image restoration with object detection with the aim to maximize detection performance over image quality. Although it performs poorly for comparatively easier benchmarks (mAP@0.15 for UAV, mAP@0.25 for Glider, mAP@0.75 for Ground), it exceeds the performance of other algorithms on difficult benchmarks (mAP @ 0.5 for Glider, mAP @ 0.9 for Ground) by almost $0.9\%$ and $0.15\%$ respectively. But as can be seen in Fig.~\ref{fig:alg_comp}, this algorithm creates many visible artifacts.

With the YOLOv$2$ architecture, the results of the submitted algorithms on the detection challenge varied greatly. Those results can be found in Supp.~Table~2. While none of the algorithms could beat the performance of the baseline for the Ground collection at mAP@0.15 and mAP@0.25, MTLab (0.41\% improvement over baseline) and UCAS-NEU (2.64\% and 0.22\% improvement over baseline) had the highest performance at mAP@0.50, mAP@0.75 and mAP@0.90 respectively, reiterating the fact that simple traditional methods like image sharpening can improve detection performance in relatively clean images. For the Glider collection, Honeywell's SRCNN-based autoencoder had the best performance at mAP@0.15 and mAP@0.25 with over 0.66\% and 0.13\% improvement over the baseline respectively, superseding their performance of 3.22\% with YOLOv$3$ (see Supp.~Fig.~2). Unlike, YOLOv$3$, YOLOv$2$ seems to be less affected by JPEG blocking artifacts that could have been enhanced due to their algorithm. For UAV however, UCAS-NEU and IMECAS-CAS had the highest performance at mAP@0.15 and mAP@0.25 with marginal improvements of 0.02\% and 0.01\% over the baselines.

\textbf{Object Classification Improvement in Video.}
Fig.~\ref{fig:alg_comp} shows a comparison of the visual effects of each of the top-performing algorithms. For most algorithms the enhancement seems to be quite subtle (as is the case of the UCAS-NEU and IMECAS-UMAC examples), when analyzing the differences with the raw image it is easier to notice the effects of their sharpening algorithms. We observe a different scenario in the images enhanced by the MTLab algorithm. The images generated with this approach present noticeable artifacts, with a larger area of the image being modified. While this behavior seemed to provide good improvement in the object detection task, it proved to be adverse for the object classification scenario in which the exact location of the object of interest is already known.

\begin{figure}[h]
    \centering
    \begin{tikzpicture}

\begin{groupplot}[
        group style={
            group size=1 by 2,
            x descriptions at=edge bottom,
            vertical sep=0pt,
            xticklabels at=edge bottom,
            /pgf/bar width=6pt
        },
        nodes near coords,
        ybar,
        every node near coord/.append style={rotate=90, anchor=west, font=\small},
        enlarge y limits={upper,value=0.25},
        ymajorgrids=true,
        grid style=dotted,
        ymax = 40,
        height=6cm,
        width=0.45*\textwidth,
        yticklabel={$\pgfmathprintnumber{\tick}\%$},
        xtick=data,
        symbolic x coords={UCAS-NEU, IMECAS-UMAC, Honeywell Intl, TUM-CAS, Sunway.AI, MTlab}
    ]
    \nextgroupplot[
        ylabel={\textbf{VGG16}},
        anchor=north west,
        xtick align=inside,
        xticklabel style={color=white},
        legend style={at={(0.5,1.15)},
              anchor=north,legend columns=-1}
    ]
        \addplot [black!60!red, fill=red!60] coordinates {(UCAS-NEU, 12.854) (IMECAS-UMAC, 12.862) (Honeywell Intl, 12.712) (TUM-CAS, 12.856) (Sunway.AI, 12.501) (MTlab, 8.429)};
        \addplot [black!60!green,fill=black!40!green] coordinates {(UCAS-NEU, 34.400) (IMECAS-UMAC, 32.229) (Honeywell Intl, 33.710) (TUM-CAS, 33.234) (Sunway.AI, 32.108) (MTlab, 16.434)};
        \addplot [black!60!blue, fill=blue!60] coordinates {(UCAS-NEU, 31.626) (IMECAS-UMAC, 29.356) (Honeywell Intl, 29.356) (TUM-CAS, 29.356) (Sunway.AI, 29.356) (MTlab, 19.035) };
        
        \coordinate (A) at (axis cs:UCAS-NEU,12.71);
        \coordinate (B) at (axis cs:UCAS-NEU,33.71);
        \coordinate (C) at (axis cs:UCAS-NEU,29.36);
        \coordinate (O1) at (rel axis cs:0,0);
        \coordinate (O2) at (rel axis cs:1,0);
        
        \draw [black!30!red,sharp plot,dashed] (A -| O1) -- (A -| O2);
        \draw [black!30!green,sharp plot, dashed] (B -| O1) -- (B -| O2);
        \draw [black!10!blue,sharp plot, dashed] (C -| O1) -- (C -| O2);
        
        \legend{UAV, Glider, Ground}
    
    \nextgroupplot[
        ylabel={\textbf{ResNet50}},
        xticklabel style={anchor= east, align=right, rotate=25, font=\small }
    ]
        \addplot [black!60!red, fill=red!60] coordinates {(UCAS-NEU, 14.8116) (IMECAS-UMAC, 14.5814) (Honeywell Intl, 15.053) (TUM-CAS, 14.9755) (Sunway.AI, 14.4628) (MTlab, 5.8148)};
        \addplot [black!60!green,fill=black!40!green] coordinates {(UCAS-NEU, 29.0675) (IMECAS-UMAC, 31.5615) (Honeywell Intl, 28.4814) (TUM-CAS, 28.621) (Sunway.AI, 31.0327) (MTlab, 8.4467)};
        \addplot [black!60!blue, fill=blue!60] coordinates {(UCAS-NEU, 39.3187) (IMECAS-UMAC, 38.6936) (Honeywell Intl, 38.6936) (TUM-CAS, 38.6936) (Sunway.AI, 38.6936) (MTlab, 12.7366) };
        
        \coordinate (A) at (axis cs:UCAS-NEU,15.05);
        \coordinate (B) at (axis cs:UCAS-NEU,28.48);
        \coordinate (C) at (axis cs:UCAS-NEU,38.69);
        \coordinate (O1) at (rel axis cs:0,0);
        \coordinate (O2) at (rel axis cs:1,0);
        
        \draw [black!30!red,sharp plot,dashed] (A -| O1) -- (A -| O2);
        \draw [black!30!green,sharp plot, dashed] (B -| O1) -- (B -| O2);
        \draw [black!10!blue,sharp plot, dashed] (C -| O1) -- (C -| O2);
    \end{groupplot}

\end{tikzpicture}\vspace{-0.5em}
    \caption{Best performing submissions per team for the object classification task across different classifiers. The dashed lines denote the baseline results for each given dataset.\vspace{-1em}}
    \label{fig:classif_top_results}
\end{figure}
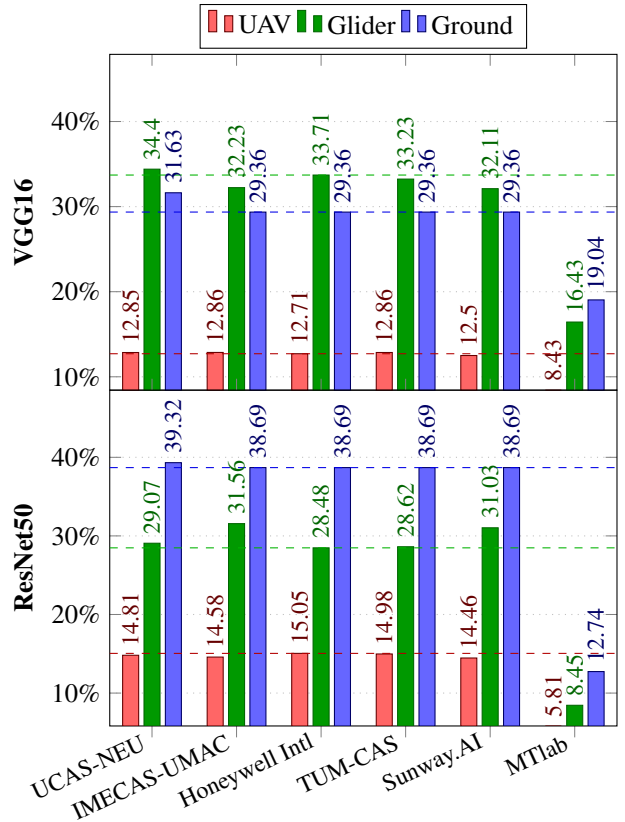

It is important to note that even though the UAV collection is quite challenging, with a baseline performance of $12.71\%$  for the VGG16 network, it tended to be the collection for which most of the evaluated algorithms presented some kind of improvement over the baseline (more detailed information on the performance of all the submitted algorithms can be found in Supp.~Tables 3-7). The highest improvement, however, was ultimately low ($0.15\%$ improvement). The intuition behind this is that even though the enhancement techniques employed by all of the algorithms were quite varied, given the high diversity of optical aberrations in the UAV collection, a good portion of the enhancement methods do not correct for all of the degradation types present in these videos. 
As for the Glider Collection, only one of the methods was able to beat the baseline classification performance by $0.689\%$ in that case. We observe similar results for the Ground collection, where only two algorithms were able to improve upon the baseline classification performance for the VGG16 network. Interestingly, a majority of the algorithms skipped the processing for the videos from this collection, considering them to be of sufficient quality by default. Nevertheless, the highest performance improvement was achieved for the Ground Collection, with the top-performing algorithm for this set providing a $4.25\%$ improvement over the baseline.  

Fig.~\ref{fig:classif_top_results} shows the top algorithm for each team in all three collections, as well as how they compare with their respective baselines for two different classifiers (VGG16 and ResNet50). When comparing the performance of the algorithms across different classifiers we observe interesting results. While one would expect a classification-driven enhancement algorithm to uniformly enhance the image ``quality'' for any given classifier (either by emphasizing the features most useful for classification or my suppressing any adversarial information in the image), we observed that the performance of most enhancement algorithms did not follow the same patterns across multiple classifiers. Some algorithms were indeed able to consistently improve the classification performance of the images for a given dataset across both classifiers (namely the UCAS-NEU algorithm was able to improve the classification performance of the images in the Glider and Ground Collections), however this behavior was not the norm. We noticed a considerable improvement on the performance of the enhanced images belonging to the Glider Collection when evaluating them using the ResNet50 classifier. Only one of the methods (UCAS-NEU) was able to obtain an improvement over the baseline performance when evaluated with the VGG16 classifier, whereas most of the algorithms were able to obtain an improvement over the baseline when evaluated with ResNet50. A similar disparity is present in the results for the UAV Collection, in this case all of the enhancement algorithms had a decrease in their performance when the evaluation classifier was ResNet50 (none of them were able to obtain better results than the baseline performance), while a good number of them were able to improve the classification results for the VGG16 network, hinting that while the performance of these networks is similar, the enhancement of features for one does not necessarily translate to an improvement in the features utilized by the other. An analysis related to this phenomenon was done by Palacio et. al.~\cite{DBLP:journals/corr/abs-1803-08337} where they studied the signal used by different classifiers to generate a prediction, out of the classifiers they evaluated they pointed out that ResNet50 would often use less than $10\%$ of the original input signal while VGG would extract features using most of it. As such, while one method could improve the data used by a given classifier, this data might not be used by a different classification model or the change might be perceived as adversarial noise by another.

\vspace{-1mm}
\section{Discussion}\vspace{-2mm}
The results of the challenge led to some surprises. While the restoration and enhancement algorithms submitted by the participants tended to improve the detection and classification results for the diverse imagery included in our dataset, no approach was able to improve the results by a significant margin. Moreover, some of the enhancement algorithms that improved performance (\textit{e.g.}, MT-Lab's approach) degraded the image quality, making it almost unrealistic. This provides evidence that most CNN-based detection methods rely on contextual features for prediction rather than focusing on the structure of the object itself. So what might seem like a perfect image to the detector may not seem realistic to a human observer. Add to this the complexity of varying scales, sizes, weather conditions and imaging artifacts like blur due to motion, atmospheric turbulence, mis-focus, distance, camera characteristics, etc. Simultaneously correcting the artifacts with the dual goal of improving recognition and perceptual quality of these videos is an enormous task --- and we have only begun to scratch the surface.

\bibliographystyle{elsarticle-num}
\bibliography{refs}

\end{document}


\title{\vspace{-15mm}Report on UG$^2+$ Challenge Track 1: Assessing  Algorithms to Improve Video Object Detection and Classification from Unconstrained Mobility Platforms\\\\ Supplemental Material\vspace{-10 mm}}

\maketitle

\section{Details about UG$^2+$ Object Detection Dataset}

\textbf{UAV Collection.} This collection consists of $26,105$ images with $30,051$ object-level annotations extracted from video recorded by small UAVs in both rural and urban areas. Some frames contain more than one object. Because the videos come from YouTube, they have different resolutions (from $600\times400$ to $3840\times2026$) and frame rates (from $12$ FPS to $59$ FPS). The imaging artifacts include but are not restricted to: glare/lens flare, poor image quality, occlusion, over/under exposure, camera shaking, noise, motion blur, fish eye lens distortion, and problematic weather conditions. This collection contains $31$ classes. Object classes that have been annotated include aircraft, trashcan, umbrella, wall, water tower, yurt, farm machines, swing, lighthouse, street signs, and pedestrians. 

\textbf{Glider Collection.} This collection consists of $19,152$ images with $22,390$ object-level annotations extracted from video recorded by gliders in both rural and urban areas. Some frames contain more than one object. The videos have frame rates ranging from $25$ FPS to $50$ FPS and different types of compression such as MTS, MP4 and MOV. Given the nature of this collection, the videos mostly present imagery taken from thousands of feet above ground, and contain artifacts such as: motion blur, occlusion, glare/lens flare, over/under exposure, camera shaking, noise, motion blur, and fisheye lens distortion. Furthermore, this collection contains videos with fog, clouds and rain. 


\textbf{Ground Collection.} This collection consists of $41,227$ frames with exactly one object per frame extracted from $136$ videos. It includes static objects (\textit{e.g.}, flower pots, buildings) at a wide range of distances ($30$ft, $40$ft, $50$ft, $60$ft, $70$ft, $100$ft, $150$ft, and $200$ft) with motion blur induced by using an orbital shaker to generate horizontal movement at different rotations per minute ($120$rpm, $140$rpm, $160$rpm, and $180$rpm). Other artifacts include different weather conditions (sun, clouds, rain, snow), and fisheye distortion due to different image sensors (GoPro versus Sony Bloggie). Different from UG$^2$, we omitted two classes from this dataset, namely ``Golfcart" and ``ResolutionChart," making the total number of object classes used in this dataset $19$.

\textbf{Dataset Partitions Released to Participants.} Approximately $75\%$ of the image frames from the object detection dataset are reserved for training and the rest are used for validation during training. This represents most of the data used to generate Fig.~\ref{fig:ap_validation} and Table~\ref{tab:dataset_mAP_v2} in the main text. 

\subsection{Average Precision on UG$^2 +$ Object Detection Validation}
\begin{figure}[ht]
    \centering
    \begin{subfigure}{0.45\textwidth}
        \centering
        \includegraphics[width=1\textwidth]{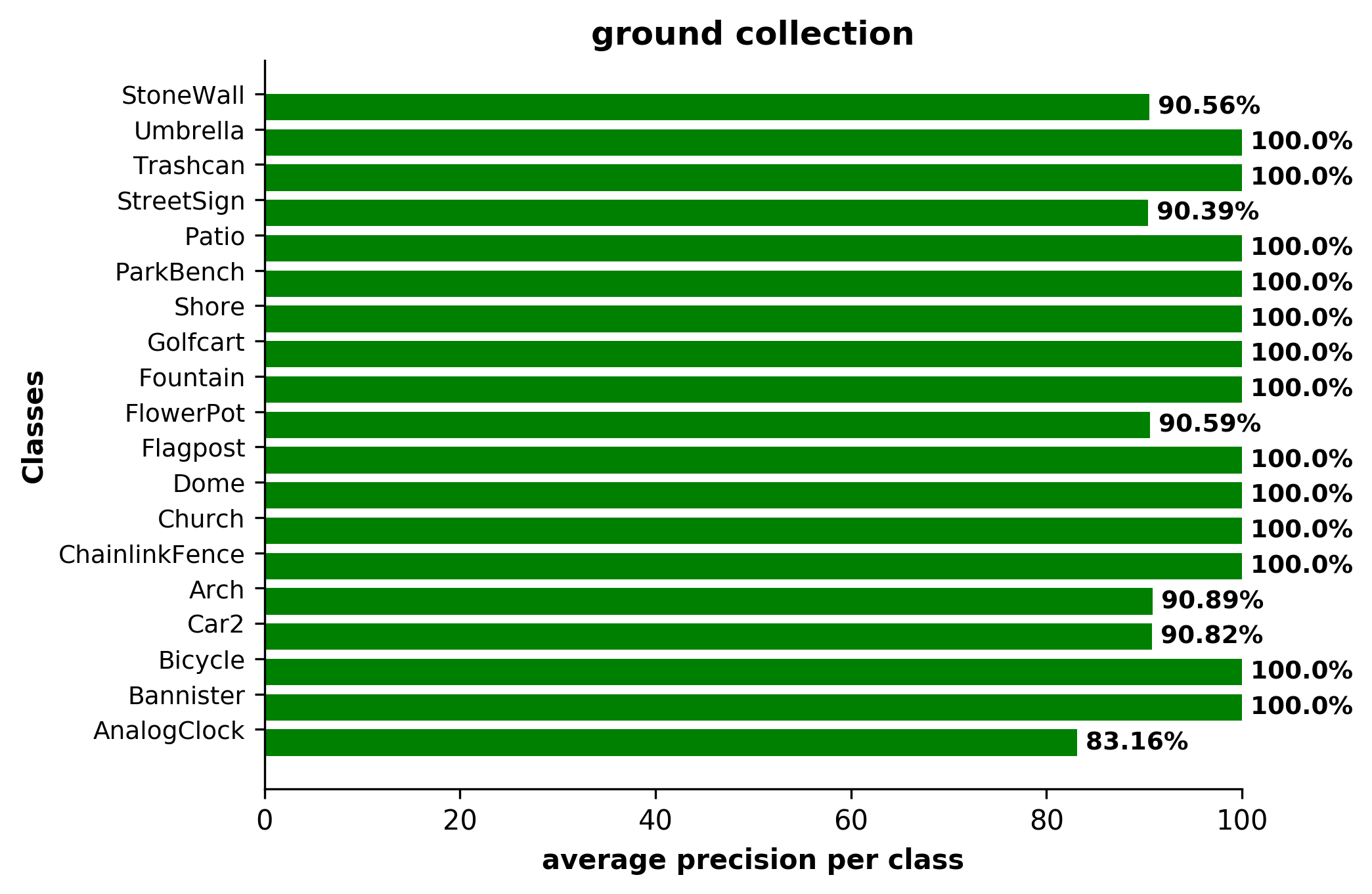}
        \vspace{2mm}
    \end{subfigure}
    \begin{subfigure}{0.45\textwidth}
        \centering
        \includegraphics[width=1\textwidth]{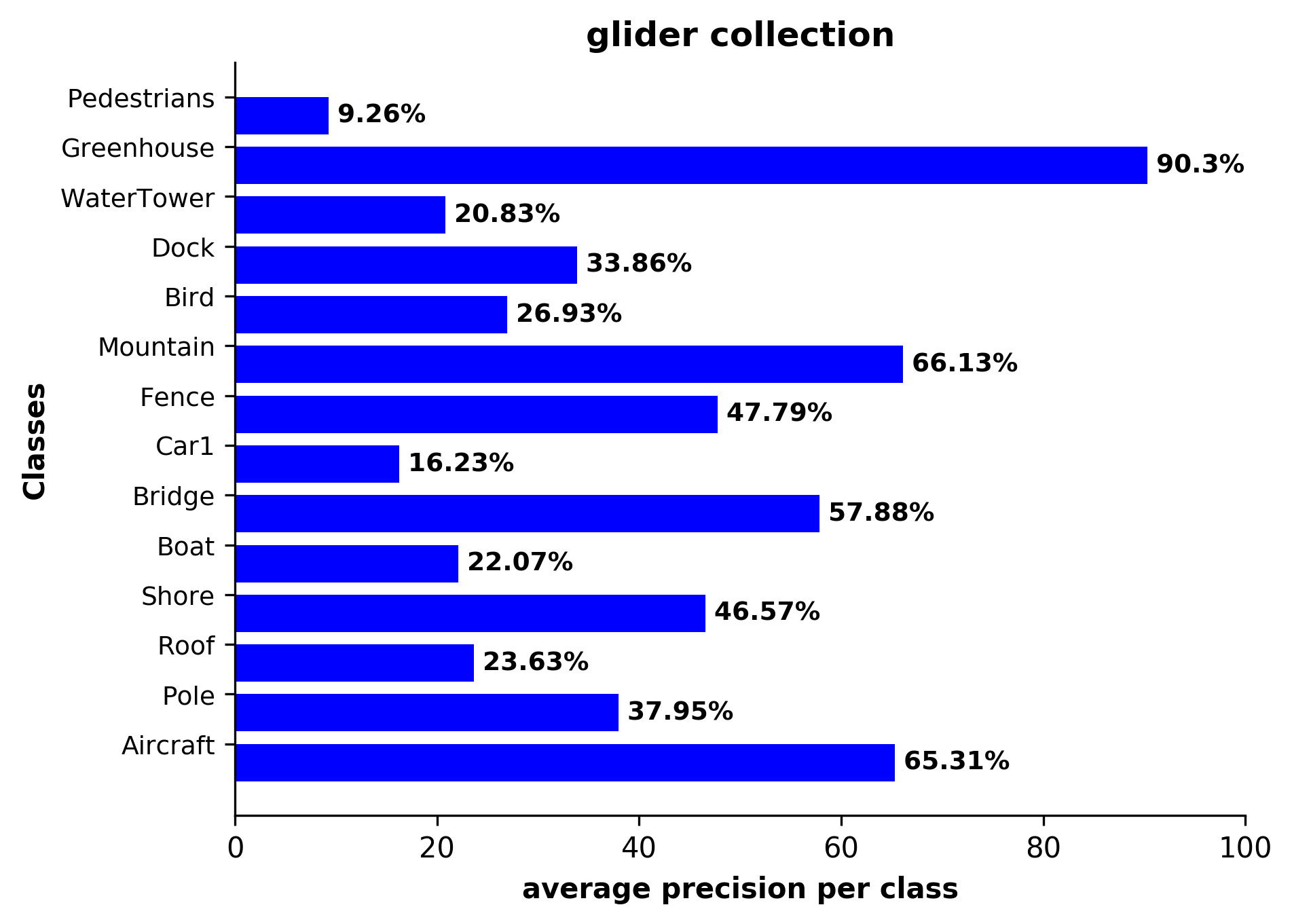}
        \vspace{2mm}
    \end{subfigure}
    \begin{subfigure}{0.45\textwidth}
        \centering
        \includegraphics[width=1\textwidth]{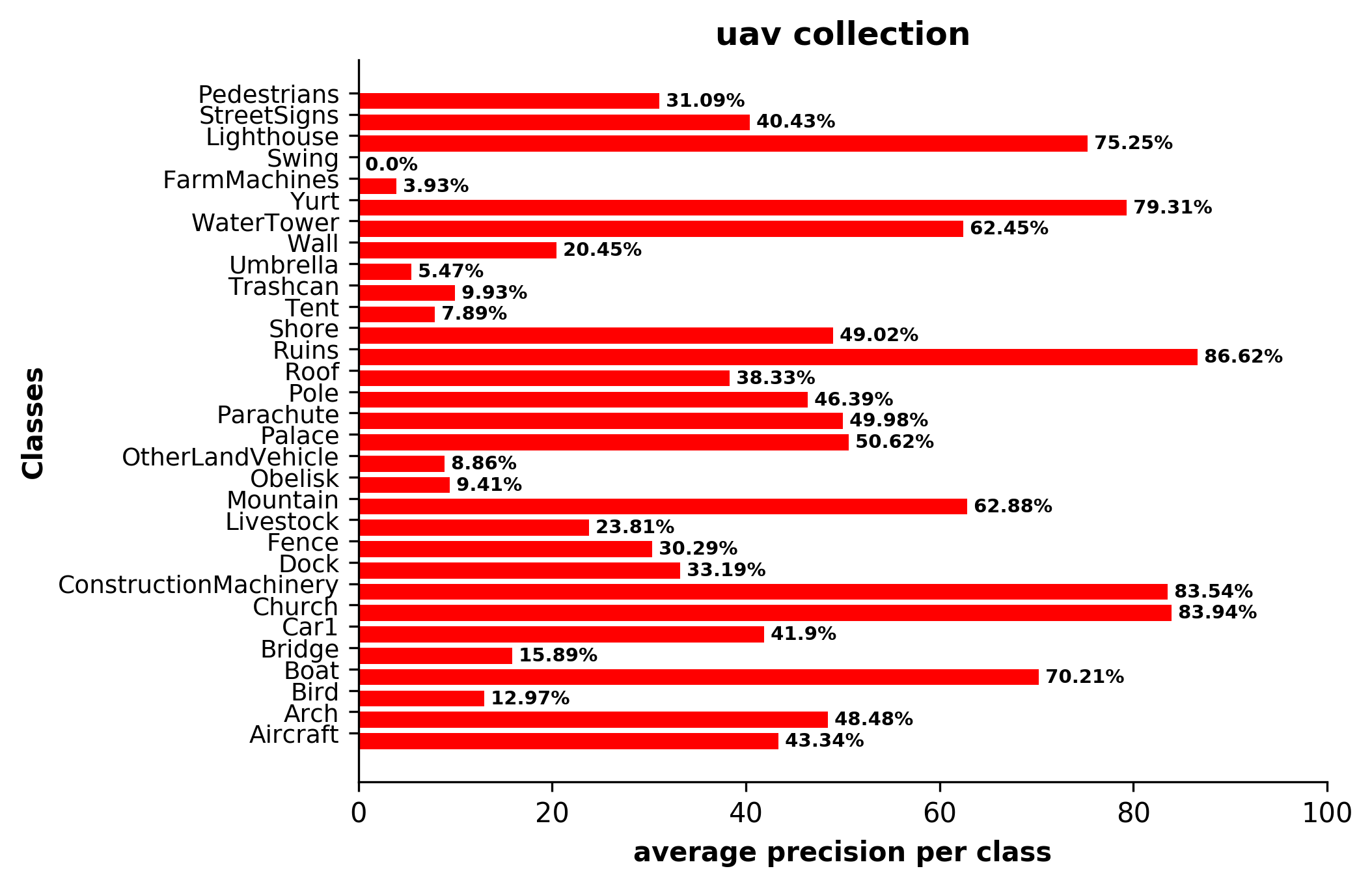}
        \vspace{2mm}
    \end{subfigure} \vspace{-2mm}
    \caption[UG$^2$ Object Detection Dataset: Average Precision per Class @ mAP$0.75$]{Average precision per class on the validation data for the three UG$^2+$ collections.}
    \label{fig:ap_validation}
    \vspace{-1mm}
\end{figure}

Fig.~\ref{fig:ap_validation} shows the average precision per class per dataset collection for mAP at $0.75$. At this operating point most of the classes in the Ground collection can be detected easily. However, some classes (\textit{e.g.}, street sign, flower pot, pedestrians, trashcan, bird) from all three collections were difficult to detect due to the scale, size and view conditions of the respective objects compared to the other classes, as well as due to the impact of imaging artifacts. We predominantly use these classes in the test dataset. Also, no two videos are the same with respect to the degree of severity of the artifacts present in them in the Glider and UAV collections, adding complexity to the detection task. 

\subsection{YOLOv2 Comparison} 
Compared to its previous variant YOLOv$2$, YOLOv$3$ has a deeper architecture and boasts of residual connections and upsampling layers. According to the authors, some salient features of YOLOv$3$ that make it better than the previous variant are its ability to make detections at three different scales and to detect smaller objects. These features are only useful, however, if the image is not corrupted by artifacts like motion blur, compression artifacts, etc. This (see Table~1 in the main paper and Table~\ref{tab:dataset_mAP_v2} here) is evident in the higher scores (31.58\%, 31.58\% and 21.53\%) it achieves for the Ground collection for mAP at $0.15$, $0.25$ and $0.50$ in comparison to YOLOv$2$'s performance (26.32\%, 26.32\% and 16.86\%). For the Glider and Ground Collections, there isn't much difference in performance for mAP at $0.15$ and $0.25$.
However, for mAP at $0.90$ for the Ground collection and mAP at $0.50$ for the Glider collection, YOLOv$2$ has better performance (3.8\% and 0.93\%) than YOLOv$3$ (0.04\% and 0.01\%). At these operating points both architectures could successfully detect larger objects like ``Arch" and ``Mountains," however YOLOv$3$ seems to be more affected by imaging artifacts due to repeated upsampling that also enlarges the artifacts and makes detection hard.

\begin{table}[]
\centering
\begin{tabular}{@{}ccccccc@{}}
\toprule
\textbf{Division} & \textbf{Collection} & \textbf{mAP @ 0.15} & \textbf{mAP @ 0.25} & \textbf{mAP @ 0.50} & \textbf{mAP @ 0.75} & \textbf{mAP @ 0.90} \\ \midrule
\multicolumn{1}{l}{\multirow{3}{*}{\textbf{Validation}}} & Ground & 100\% & 100\% & 100\% & 96.84\% & 59.34\% \\
\multicolumn{1}{l}{} & Glider & 94.91\% & 93.87\% & 87.24\% & 47.66\% & 7.56\% \\
\multicolumn{1}{l}{} & UAV & 96.94\% & 96.07\% & 91.87\% & 55.90\% & 7.61\% \\ \cmidrule(l){2-7} 
\multirow{3}{*}{\textbf{Test}} & Ground & 26.32\% & 26.32\% & 16.86\% & 11.69\% & 3.80\% \\
 & Glider & 4.94\% & 4.92\% & 0.93\% & 0\% & 0\% \\
 & UAV & 0.90\% & 0.38\% & 0.000002\% & 0\% & 0\% \\ \cmidrule(l){2-7} 
\end{tabular}
\caption{mAP scores for the UG$^2$+ Object Detection Dataset with YOLOv2 fine-tuned on UG$^2$ dataset.}
\label{tab:dataset_mAP_v2}
\end{table}

\subsection{Object Detection Evaluation Scores}
\begin{figure}[!h]
    \centering
    \includegraphics[width=0.90\linewidth]{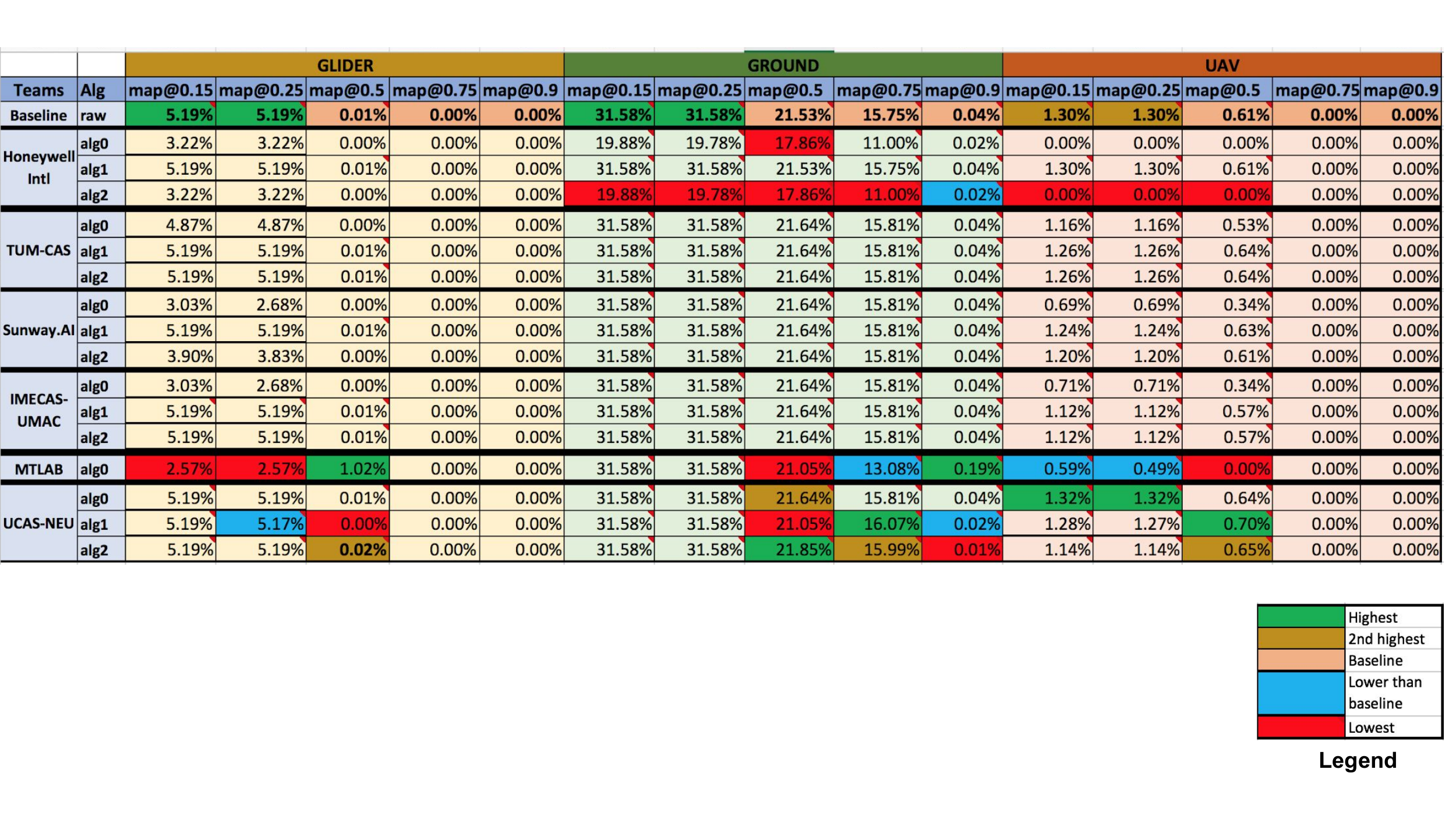}
    \caption{Complete object detection evaluation results for the participants' algorithms with YOLOv$3$ fine-tuned on UG$^2 +$ object detection dataset. The highest, $2$nd highest, and baseline algorithms' scores on the test datasets are highlighted according to the legend on the lower-right.}
    \label{fig:objdet_scores}
\end{figure}

Fig.~\ref{fig:objdet_scores} shows the complete evaluation on the object detection test dataset for the three collections for the participants' algorithm compared to the baseline with YOLOv$3$ fine-tuned on the object detection dataset. The highest, $2$nd highest, and baseline algorithms' scores are highlighted according to the legend on the lower-right. 
Table~\ref{tab:dataset_mAP_v2p} shows the result of participants' algorithm on the UG$^2+$ object detection test dataset using YOLOv2.

\begin{table}[]
\centering
\resizebox{\textwidth}{!}{%
\begin{tabular}{@{}ccccccccccccccccc@{}}
\toprule
\textbf{Collections} & \multicolumn{6}{c}{\textbf{Ground}} & \multicolumn{5}{c}{\textbf{Glider}} & \multicolumn{5}{c}{\textbf{UAV}} \\ \midrule
\textbf{Participants} & \textbf{Algorithms} & \textbf{mAP @ 0.15} & \textbf{mAP @ 0.25} & \textbf{mAP @ 0.50} & \textbf{mAP @ 0.75} & \textbf{mAP @ 0.90} & \textbf{mAP @ 0.15} & \textbf{mAP @ 0.25} & \textbf{mAP @ 0.50} & \textbf{mAP @ 0.75} & \textbf{mAP @ 0.90} & \textbf{mAP @ 0.15} & \textbf{mAP @ 0.25} & \textbf{mAP @ 0.50} & \textbf{mAP @ 0.75} & \textbf{mAP @ 0.90} \\ \midrule
\textbf{Baseline} & -- & \textbf{26.32} & \textbf{26.32} & 16.86 & 11.69 & 3.80 & 4.94 & 4.92 & \textbf{0.93} & 0 & 0 & 0.90 & 0.38 & 0 & 0 & 0 \\ \midrule
\textbf{MTLab} & alg0 & 26.31 & 26.31 & \textbf{17.27} & 5.36 & 0.45 & 5.31 & 5.04 & 0.62 & 0 & 0 & 0.84 & 0.02 & 0 & 0 & 0 \\ \midrule
\multirow{3}{*}{\textbf{Honeywell}} & alg0 & 17.38 & 17.32 & 15.15 & 5.27 & 0.01 & \textbf{5.60} & \textbf{5.05} & 0.55 & 0 & 0 & 0.09 & 0.06 & 0 & 0 & 0 \\
 & alg1 & 26.32 & 26.32 & 16.86 & 11.69 & 3.80 & 4.94 & 4.92 & 0.93 & 0 & 0 & 0.90 & 0.38 & 0 & 0 & 0 \\
 & alg2 & 17.38 & 17.32 & 15.15 & 5.27 & 0.01 & \textbf{5.60} & \textbf{5.05} & 0.55 & 0 & 0 & 0.09 & 0.06 & 0 & 0 & 0 \\ \midrule
\multirow{3}{*}{\textbf{UCAS-NEU}} & alg0 & 26.32 & 26.32 & 16.85 & 11.71 & 3.80 & 4.96 & 4.93 & 0.93 & 0 & 0 & 0.90 & 0.38 & 0 & 0 & 0 \\
 & alg1 & 26.32 & 26.32 & 16.85 & \textbf{14.33} & 3.76 & 4.58 & 4.47 & 0.25 & 0 & 0 & \textbf{0.92} & 0.09 & 0 & 0 & 0 \\
 & alg2 & 26.32 & 26.32 & 17.06 & 12.27 & \textbf{4.02} & 4.94 & 4.87 & 1.07 & 0 & 0 & 0.90 & \textbf{0.39} & 0 & 0 & 0 \\ \midrule
\multirow{3}{*}{\textbf{TUM-CAS}} & alg0 & 26.32 & 26.32 & 16.85 & 11.71 & 3.80 & 4.97 & 4.94 & 0.93 & 0 & 0 & 0.90 & 0.10 & 0 & 0 & 0 \\
 & alg1 & 26.32 & 26.32 & 16.85 & 11.71 & 3.80 & 4.91 & 4.88 & 0.92 & 0 & 0 & 0.90 & 0.38 & 0 & 0 & 0 \\
 & alg2 & 26.32 & 26.32 & 16.85 & 11.71 & 3.80 & 4.96 & 4.93 & 0.93 & 0 & 0 & 0.90 & 0.38 & 0 & 0 & 0 \\ \midrule
\multicolumn{1}{l}{\multirow{3}{*}{\textbf{IMECAS-UCAS}}} & alg0 & 26.32 & 26.32 & 16.85 & 11.71 & 3.80 & 4.66 & 4.48 & 0.17 & 0 & 0 & \textbf{0.92} & 0.09 & 0 & 0 & 0 \\
\multicolumn{1}{l}{} & alg1 & 26.32 & 26.32 & 16.85 & 11.71 & 3.80 & 5.00 & 4.98 & 0.67 & 0 & 0 & 0.90 & \textbf{0.39} & 0 & 0 & 0 \\
\multicolumn{1}{l}{} & alg2 & 26.32 & 26.32 & 16.85 & 11.71 & 3.80 & 5.00 & 4.98 & 0.67 & 0 & 0 & 0.90 & \textbf{0.39} & 0 & 0 & 0 \\ \midrule
\multirow{3}{*}{\textbf{SUNWAY.AI}} & alg0 & 26.32 & 26.32 & 16.85 & 11.71 & 3.80 & 4.66 & 4.48 & 0.17 & 0 & 0 & 0.91 & 0.09 & 0 & 0 & 0 \\
 & alg1 & 26.32 & 26.32 & 16.85 & 11.71 & 3.80 & 5.00 & 4.98 & 0.67 & 0 & 0 & 0.91 & 0.09 & 0 & 0 & 0 \\
 & alg2 & 26.32 & 26.32 & 16.85 & 11.71 & 3.80 & 3.73 & 3.73 & 0.15 & 0 & 0 & 0.91 & 0.90 & 0 & 0 & 0 \\ \bottomrule
\end{tabular}%
}
\caption{mAP scores for the UG$^2$+ Object Detection Validation Dataset (in percentage) with YOLOv2 fine-tuned on the UG$^2$ dataset. }
\label{tab:dataset_mAP_v2p}
\end{table}

\section{Object Classification}
Tables~\ref{tab:classif:all_results_vgg16}-\ref{tab:classif:all_results_NASNetMobile} show the average LRAP of each submitted algorithms on the test set using VGG16, ResNet50, DenseNet201, MobileNetV2 and NASNetMobile classifiers. Bold entries denote the best performing algorithm for a given collection. Underlined entries denote the algorithm had an average LRAP higher than the baseline performance for that collection.


\begin{table}[!htb]
\setlength\tabcolsep{4.5pt}
    \begin{minipage}{.5\linewidth}
      \caption{VGG16}\label{tab:classif:all_results_vgg16}
      \centering
        \begin{tabular}{|c|c|c|c|c|}
            \hline
            \textbf{Team} & \textbf{Alg} & \textbf{UAV} & \textbf{Glider} & \textbf{Ground} \\ \hline
            \multicolumn{2}{|c|}{Baseline} & 12.71\% & 33.71\% & 29.36\% \\ \hline
            Honeywell Intl & alg0 & 5.42\% & 23.18\% & 12.42\% \\
            Honeywell Intl & alg1 & 12.71\% & 33.71\% & 29.36\% \\
            Honeywell Intl & alg2 & 5.42\% & 23.18\% & 12.42\% \\
            IMECAS-UMAC & alg0 & {\ul \textbf{12.86\%}} & 28.43\% & 29.36\% \\
            IMECAS-UMAC & alg1 & 12.6831\% & 32.23\% & 29.36\% \\
            IMECAS-UMAC & alg2 & \textbf{12.86\%} & 32.23\% & 29.36\% \\
            MTlab & alg0 & 8.43\% & 16.43\% & 19.035\% \\
            SunwayAI & alg0 & 12.48\% & 28.45\% & 29.36\% \\
            SunwayAI & alg1 & 12.50\% & 32.11\% & 29.36\% \\
            SunwayAI & alg2 & 12.47\% & 30.01\% & 29.36\% \\
            TUM-CAS & alg0 & 12.77\% & 33.19\% & 29.36\% \\
            TUM-CAS & alg1 & 12.83\% & 33.13\% & 29.36\% \\
            TUM-CAS & alg2 & 12.86\% & 33.24\% & 29.36\% \\
            UCAS-NEU & alg0 & 12.71\% & 33.71\% & 29.36\% \\
            UCAS-NEU & alg1 & 12.71\% & 33.44\% & \textbf{33.61\%} \\
            UCAS-NEU & alg2 & 12.85\% & \textbf{34.40\%} & 31.63\% \\ \hline
        \end{tabular}
    \end{minipage}%
    \begin{minipage}{.5\linewidth}
      \centering
        \caption{ResNet50}\label{tab:classif:all_results_resnet}
        \begin{tabular}{|c|c|c|c|c|}
            \hline
            \textbf{Team} & \textbf{Alg} & \textbf{UAV} & \textbf{Glider} & \textbf{Ground} \\ \hline
            \multicolumn{2}{|c|}{Baseline} & 15.05\% & 28.48\% & 38.69\% \\ \hline
            Honeywell Intl & alg0 & 6.22\% & 20.97\% & 9.21\% \\
            Honeywell Intl & alg1 & 15.05\% & 28.48\% & 38.69\% \\
            Honeywell Intl & alg2 & 6.21\% & 20.97\% & 9.21\% \\
            IMECAS-UMAC & alg0 & 14.81\% & 28.91\% & 38.69\% \\
            IMECAS-UMAC & alg1 & 14.58\% & \textbf{31.56\%} & 38.69\% \\
            IMECAS-UMAC & alg2 & 14.82\% & \textbf{31.56\%} & 38.69\% \\
            MTlab & alg0 & 5.81\% & 8.45\% & 12.74\% \\
            SunwayAI & alg0 & 14.48\% & 28.89\% & 38.69\% \\
            SunwayAI & alg1 & 14.47\% & 31.03\% & 38.69\% \\
            SunwayAI & alg2 & 14.46\% & 29.43\% & 38.69\% \\
            TUM-CAS & alg0 & 14.50\% & 28.39\% & 38.69\% \\
            TUM-CAS & alg1 & 14.98\% & 28.62\% & 38.69\% \\
            TUM-CAS & alg2 & 14.98\% & 28.90\% & 38.69\% \\
            UCAS-NEU & alg0 & 15.05\% & 28.48\% & 38.69\% \\
            UCAS-NEU & alg1 & 14.57\% & 27.82\% & \textbf{39.82\%} \\
            UCAS-NEU & alg2 & 14.81\% & 29.07\% & 39.32\% \\ \hline
        \end{tabular}
    \end{minipage} 
    
\end{table}

\begin{table}[!htb]
\setlength\tabcolsep{4.5pt}
    \begin{minipage}{.5\linewidth}
      \caption{DenseNet201}\label{tab:classif:all_results_densenet}
      \centering
       \begin{tabular}{|c|c|c|c|c|}
        \hline
        \textbf{Team} & \textbf{Alg} & \textbf{UAV} & \textbf{Glider} & \textbf{Ground} \\ \hline
        \multicolumn{2}{|c|}{Baseline} & 2.06\% & 1.00\% & 3.84\% \\ \hline
        Honeywell Intl & alg0 & \textbf{3.99\%} & \textbf{3.46\%} & \textbf{5.38\%} \\
        Honeywell Intl & alg1 & 2.06\% & 1.00\% & 3.84\% \\
        Honeywell Intl & alg2 & \textbf{3.99\%} & \textbf{3.46\%} & \textbf{5.38\%} \\
        IMECAS-UMAC & alg0 & 2.00\% & 0.58\% & 3.84\% \\
        IMECAS-UMAC & alg1 & 1.91\% & 0.74\% & 3.84\% \\
        IMECAS-UMAC & alg2 & 2.00\% & 0.74\% & 3.84\% \\
        MTlab & alg0 & 3.90\% & 1.45\% & 4.6469\% \\
        SunwayAI & alg0 & 1.85\% & 0.57\% & 3.84\% \\
        SunwayAI & alg1 & 1.87\% & 0.72\% & 3.84\% \\
        SunwayAI & alg2 & 1.87\% & 0.54\% & 3.84\% \\
        TUM-CAS & alg0 & 2.04\% & 0.88\% & 3.84\% \\
        TUM-CAS & alg1 & 2.03\% & 0.89\% & 3.84\% \\
        TUM-CAS & alg2 & 2.03\% & 0.92\% & 3.84\% \\
        UCAS-NEU & alg0 & 2.06\% & 1.00\% & 3.84\% \\
        UCAS-NEU & alg1 & 1.97\% & 0.83\% & 4.12\% \\
        UCAS-NEU & alg2 & 2.03\% & 0.92\% & 3.70\% \\ \hline
        \end{tabular}
    \end{minipage}%
    \begin{minipage}{.5\linewidth}
      \centering
        \caption{MobileNetV2}\label{tab:classif:all_results_mobilenet}
        \begin{tabular}{|c|c|c|c|c|}
            \hline
            \textbf{Team} & \textbf{Alg} & \textbf{UAV} & \textbf{Glider} & \textbf{Ground} \\ \hline
            \multicolumn{2}{|c|}{Baseline} & 1.81\% & 1.25\% & 5.37\% \\ \hline
            Honeywell Intl & alg0 & \textbf{3.80\%} & \textbf{6.53\%} & 3.92\% \\
            Honeywell Intl & alg1 & 1.81\% & 1.25\% & 5.37\% \\
            Honeywell Intl & alg2 & 3.80\% & 6.53\% & 3.92\% \\
            IMECAS-UMAC & alg0 & 1.90\% & 1.56\% & 5.37\% \\
            IMECAS-UMAC & alg1 & 1.74\% & 1.30\% & 5.367\% \\
            IMECAS-UMAC & alg2 & 1.74\% & 1.30\% & 5.37\% \\
            MTlab & alg0 & 1.70\% & 0.79\% & 6.13\% \\
            SunwayAI & alg0 & 1.81\% & 1.57\% & 5.37\% \\
            SunwayAI & alg1 & 1.80\% & 1.28\% & 5.37\% \\
            SunwayAI & alg2 & 1.80\% & 1.60\% & 5.37\% \\
            TUM-CAS & alg0 & 1.80\% & 1.23\% & 5.37\% \\
            TUM-CAS & alg1 & 1.81\% & 1.31\% & 5.37\% \\
            TUM-CAS & alg2 & 1.81\% & 1.28\% & 5.37\% \\
            UCAS-NEU & alg0 & 1.81\% & 1.25\% & 5.37\% \\
            UCAS-NEU & alg1 & 1.89\% & 2.35\% & \textbf{6.48\%} \\
            UCAS-NEU & alg2 & 1.74\% & 1.13\% & 4.76\% \\ \hline
        \end{tabular}
    \end{minipage} 
    
\end{table}

\begin{table}[!htb]
\setlength\tabcolsep{4.5pt}
    \begin{minipage}{.5\linewidth}
      \caption{NASNetMobile}\label{tab:classif:all_results_NASNetMobile}
      \centering
       \begin{tabular}{|c|c|c|c|c|}
            \hline
            \textbf{Team} & \textbf{Alg} & \textbf{UAV} & \textbf{Glider} & \textbf{Ground} \\ \hline
            \multicolumn{2}{|c|}{Baseline} & 2.04\% & 0.46\% & 0.51\% \\ \hline
            Honeywell Intl & alg0 & \textbf{2.34\%} & \textbf{1.19\%} & \textbf{1.98\%} \\
            Honeywell Intl & alg1 & 2.04\% & 0.46\% & 0.51\% \\
            Honeywell Intl & alg2 & \textbf{2.34\%} & \textbf{1.19\%} & \textbf{1.98\%} \\
            IMECAS-UMAC & alg0 & 1.51\% & 0.94\% & 0.51\% \\
            IMECAS-UMAC & alg1 & 1.93\% & 0.56\% & 0.51\% \\
            IMECAS-UMAC & alg2 & 1.93\% & 0.56\% & 0.51\% \\
            MTlab & alg0 & 1.75\% & 0.38\% & 0.76\% \\
            SunwayAI & alg0 & 1.52\% & 0.94\% & 0.51\% \\
            SunwayAI & alg1 & 1.52\% & 0.57\% & 0.51\% \\
            SunwayAI & alg2 & 1.52\% & 0.73\% & 0.51\% \\
            TUM-CAS & alg0 & 2.00\% & 0.66\% & 0.51\% \\
            TUM-CAS & alg1 & 2.04\% & 0.55\% & 0.51\% \\
            TUM-CAS & alg2 & 2.04\% & 0.49\% & 0.51\% \\
            UCAS-NEU & alg0 & 2.04\% & 0.46\% & 0.51\% \\
            UCAS-NEU & alg1 & 1.50\% & 0.50\% & 0.54\% \\
            UCAS-NEU & alg2 & 1.93\% & 0.46\% & 0.50\% \\ \hline
        \end{tabular}
    \end{minipage}%
    \begin{minipage}{.5\linewidth}
      \centering
        \caption{Comparison of the VGG16 classification performance on the Glider Collection. 
        }
        \setlength\tabcolsep{4.5pt}
        \begin{tabular}{|l|l|l|}
            \hline
            \textbf{Class} & \textbf{LRAP\_train} & \textbf{LRAP\_test} \\ \hline
            Aircraft & 40.70\% & 72.45\% \\ \hline
            Mountain & 71.27\% & 60.90\% \\ \hline
            Palace & 6.18\% & 51.78\% \\ \hline
            Roof & 1.87\% & 7.37\% \\ \hline
            Shore & 22.30\% & 28.07\% \\ \hline \hline
            \textbf{TOTAL AVG} & 28.46\% & 44.12\% \\ \hline
        \end{tabular}
    \end{minipage} 
\end{table}